\documentclass[10pt,twocolumn,letterpaper]{article}

\usepackage{cvpr}              % To produce the CAMERA-READY version
\usepackage[dvipsnames]{xcolor}

\definecolor{cvprblue}{rgb}{0.21,0.49,0.74}
\usepackage[pagebackref,breaklinks,colorlinks,citecolor=cvprblue]{hyperref}
\usepackage[capitalize]{cleveref}
\usepackage{kotex}
\usepackage{multirow}
\usepackage{graphicx}
\usepackage{amsmath}
\usepackage{amssymb}
\usepackage{booktabs}
\usepackage{tikz}
\usepackage{dashrule}
\usepackage{mdframed}
\usepackage{tabularx}
\usepackage[title]{appendix}
\usepackage{multibib}
\usepackage[accsupp]{axessibility}

\crefname{section}{Sec.}{Secs.}
\Crefname{section}{Section}{Sections}
\Crefname{table}{Table}{Tables}
\crefname{table}{Tab.}{Tabs.}

\def\paperID{9060} % *** Enter the Paper ID here
\def\confName{CVPR}
\def\confYear{2024}
\newcites{supple}{Referencesss}

\begin{document}

\title{Selectively Informative Description can Reduce Undesired Embedding Entanglements in Text-to-Image Personalization}
%\title{Selectively Informative Description for Reducing Undesired Embedding Entanglements in Text-to-Image Personalization}

%%%%%%%%% AUTHORS - PLEASE UPDATE
\author{Jimyeong Kim$^1$ \qquad Jungwon Park$^1$ \qquad Wonjong Rhee$^{1,2,3}$\\
Department of Intelligence and Information$^1$ \& IPAI$^2$ \& RICS$^3$, Seoul National University\\
% $^1$Department of Intelligence and Information, Seoul National University\\
% $^2$Interdisciplinary Program in Artificial Intelligence (IPAI), Seoul National University\\
% $^3$Research Institute for Convergence Science, Seoul National University\\
{\tt\small \{wlaud1001, quoded97, wrhee\}@snu.ac.kr}
% \and
}

\maketitle

\begin{abstract}
In text-to-image personalization, a timely and crucial challenge is the tendency of generated images overfitting to the biases present in the reference images.
We initiate our study with a comprehensive categorization of the biases into background, nearby-object, tied-object, substance (in style re-contextualization), and pose biases. 
These biases manifest in the generated images due to their entanglement into the subject embedding. This undesired embedding entanglement not only results in the reflection of biases from the reference images into the generated images but also notably diminishes the alignment of the generated images with the given generation prompt.
To address this challenge, we propose SID~(Selectively Informative Description), a text description strategy that deviates from the prevalent approach of only characterizing the subject’s class identification. 
SID is generated utilizing multimodal GPT-4 and can be seamlessly integrated into optimization-based models.
We present comprehensive experimental results along with analyses of cross-attention maps, subject-alignment, non-subject-disentanglement, and text-alignment.
\end{abstract}

   % In text-to-image personalization, diffusion models excel at generating images of subjects based on a few reference images, capturing the subject's identity in novel contexts. However, these models suffer from undesired embedding entanglement, where undesired contents from the reference images unintentionally infiltrate the subject's embedding, leading to unnatural or misaligned surroundings in the generated images. We identify five key biases exacerbating undesired embedding entanglement: background, nearby-object, tied-object, substance (in style re-contextualization), and pose biases. To address these biases, we introduce SID (Selectively Informative Description), a text description that encapsulates both the subject's and any undesirable objects' class identities, with informative specifications exclusively reserved for the latter. SID is generated using multimodal GPT-4 and seamlessly integrated into per-subject optimization. Integrating SID consistently enhances subject and text alignments, reducing undesired embedding entanglement in widely used personalization datasets, highly biased datasets, and even when there's only a single reference image. Additionally, we show that SID outperforms other simple alternatives, such as employing negative prompts in inference or adopting segmentation masks in per-subject optimization.
\section{Introduction}
\begin{figure*}[t]
    \centering
\includegraphics[width=0.91\textwidth]{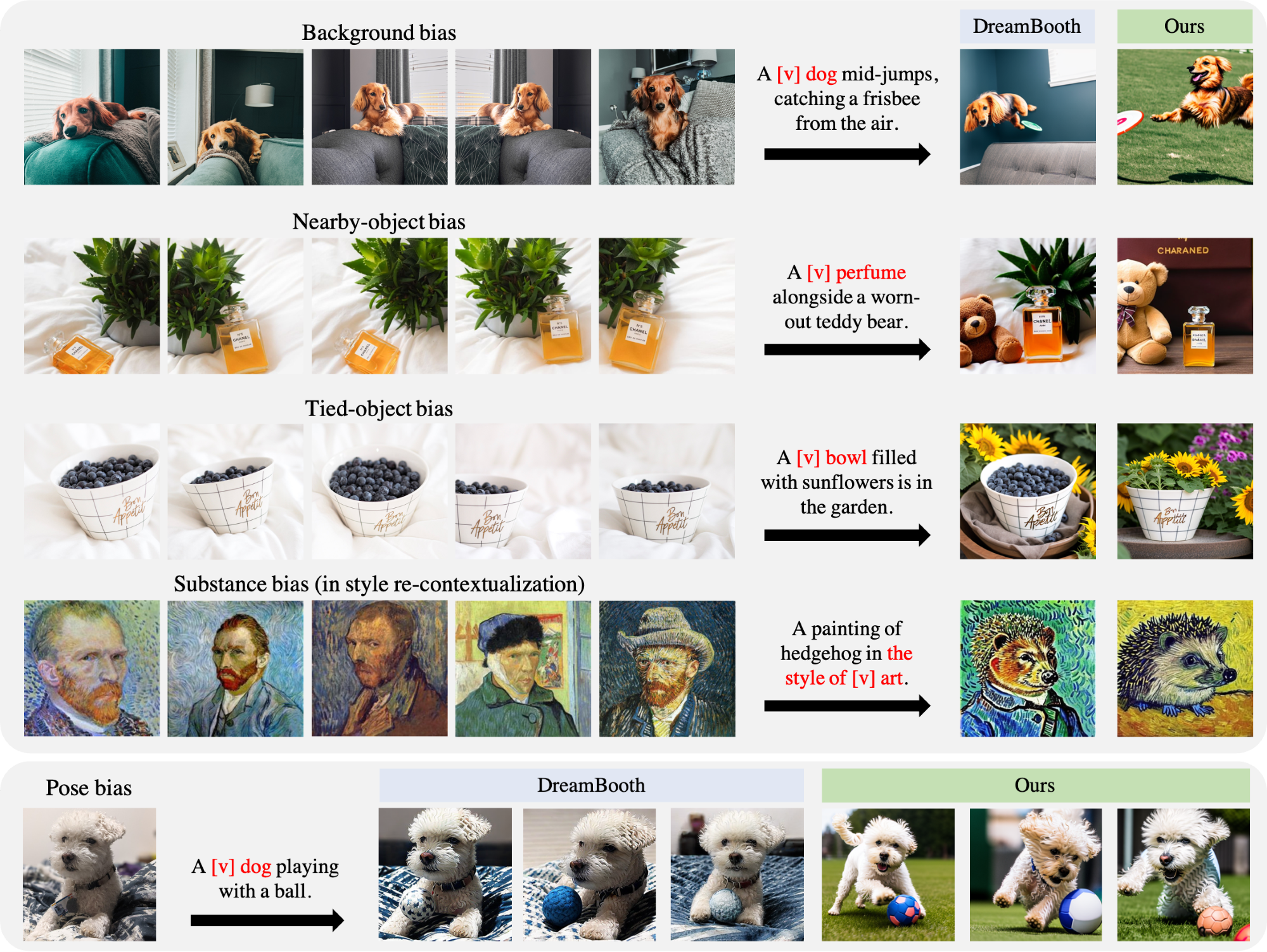}
\captionof{figure}{\textbf{Five key biases -- background, nearby-object, tied-object, substance (in style re-contextualization), and pose biases.} 
The first four rows depict scenarios with multiple reference images, while the last row illustrates a single reference image scenario. 
The pose bias is particularly prone to manifest in scenarios involving a single reference image, although it can also occur when multiple reference images depict the subject in poses that are similar.
In the generation prompt, the subject of interest is highlighted in red.
The integration of our method into DreamBooth~\cite{ruiz2023dreambooth} effectively resolves embedding entanglements associated with the five key biases (rightmost column). 
}
\vspace{-0.5cm}
% \captionof{figure}{{Five key biases in a few reference images exacerbate undesired embedding entanglements (left and middle). The integration of our method into DreamBooth~\cite{ruiz2023dreambooth} effectively resolves embedding entanglements associated with background, nearby-object, tied-object, substance (in style re-contextualization), and pose biases (right). Our approach is dedicated to preserving only the subject of interest (highlighted in red), resulting in precise re-contextualized images within their natural surroundings. The first four rows depict scenarios with multiple reference images, while the last row illustrates a single-reference image scenario.}}
\label{fig:main_bias}
\end{figure*}

Text-to-image diffusion models~\cite{nichol2021glide, saharia2022photorealistic, rombach2022high, ramesh2022hierarchical, chefer2023attend, feng2022training, phung2023grounded, liu2022compositional, lee2023aligning, black2023training} have demonstrated remarkable generation capabilities that align with textual descriptions across various applications. One such application is text-to-image personalization~\cite{gal2022image, ruiz2023dreambooth, kumari2023multi}, where models are tailored to generate novel renditions of subjects described in a few reference images. Recent works, such as DreamBooth~\cite{ruiz2023dreambooth} or Custom Diffusion~\cite{kumari2023multi}, stand out for their exceptional personalized generation results. They achieve this by fine-tuning a pre-trained text-to-image diffusion model with a small number of reference images using a text description in the format of ``a [v] [class name]'' or ``photo of a [v] [class name].'' In this context, [v] represents the subject’s unique identifier, which uses a rare token with minimal semantic significance, while the class descriptor [class name] represents a coarse category for the subject.

Recent developments in text-to-image personalization can be categorized into optimization-based approaches~\cite{gal2022image, ruiz2023dreambooth, kumari2023multi, han2023svdiff, tewel2023key} and encoder-based approaches~\cite{wei2023elite, shi2023instantbooth, li2023blip, jia2023taming}. 
For both types of approaches, a problem typically addressed as `overfitting' has emerged~\cite{avrahami2023break, ruiz2023dreambooth, tewel2023key, han2023svdiff, li2023blip, jia2023taming}. It is a phenomenon where objects, other than the subject of interest, in the reference images affect the generated image in an undesired way. 
For the sake of clarity, we refer to such an entity as `non-subject' or `undesired object'.
The cause of this phenomenon is closely related to the embedding of the identifier token [v] for optimization-based approaches and the embedding of the subject for encoder-based models, where the information related to any non-subject seeps into the embeddings. Rather than referring to it as overfitting, we formally address this phenomenon as \emph{undesired embedding entanglement}.

The phenomenon of undesired embedding entanglement has been noticed, but it has been only superficially understood so far. A potential explanation for the limited understanding is the rarity of this phenomenon in straightforward scenarios. Instances where reference images encompass diverse subject views and minimal undesired objects are much less prone to this phenomenon.
% Instances where reference images encompass diverse contexts and subject views, or contain minimal undesired objects, are much less prone to this phenomenon.
% Instances where reference images encompass the subject in diverse contexts or views, or contain minimal undesired objects, are much less prone to this phenomenon.
% Instances where reference images encompass diverse subject views, \textcolor{orange}{contain minimal undesired objects}, and involve simple generation prompts are much less prone to this phenomenon.
However, such an ideal scenario is highly unlikely in real-world applications. Therefore, it is crucial to comprehensively understand the biases and develop counter methods to address them effectively.
In an effort to gain a more thorough understanding of the prevalent biases contributing to these entanglements, we conducted exhaustive paper reviews followed by extensive experiments utilizing state-of-the-art models. The resulting identification of the five most commonly encountered biases is presented in \cref{fig:main_bias}. 
While we have chosen DreamBooth~\cite{ruiz2023dreambooth} for generating the examples in \cref{fig:main_bias}, additional examples for Custom Diffusion~\cite{kumari2023multi}, SVDiff~\cite{han2023svdiff}, ELITE~\cite{wei2023elite} and BLIP-Diffusion~\cite{li2023blip} will be provided later. 

In prior studies, several models~\cite{jia2023taming, avrahami2023break, wei2023elite, shi2023instantbooth, li2023blip} have employed segmentation masks. However, depending exclusively on segmentation masks comes with limitations, as we will demonstrate with additional examples in subsequent sections. Notably, these limitations become apparent in the cases of substance, tied-object, and pose biases. 
To tackle all of the key biases, we introduce SID (Selectively Informative Description) that requires only modifications in the train text descriptions of the reference images. The idea is very simple. 
Starting with an optimization-based model, we diverge from the prevalent approach of only characterizing the subject's class identification as observed in the existing works. Our approach involves integrating informative specifications of the ``undesired objects'' into the train descriptions. 
This modification significantly diminishes the probability of undesired entanglement between the subject embedding [v] and the non-subject information in the reference images.
As the text-to-image diffusion models are trained for alignments, adding text descriptions that match non-subjects can greatly help the non-subject parts in the reference images to align with the corresponding %added
text descriptions.
This prevents the undesired objects from accidentally becoming aligned with [v].
In contrast, such an accidental alignment can quite easily occur in the absence of the matching text descriptions.
% This is because text-to-image diffusion models are trained for alignments.
% Therefore, adding text descriptions that match non-subjects can greatly help the corresponding non-subject parts in the reference images to align with the added text descriptions.
% This prevents non-subjects from accidentally becoming aligned with [v].
% In contrast, such an accidental alignment can quite easily occur in the absence of the matching text descriptions.
% 243012 Orig:
% This modification significantly diminishes the probability of entanglement between the subject embedding [v] and the undesired objects in the reference images, because such non-subject information almost always aligns with the corresponding informative specifications in the train description instead. 
% 
%
% While infor for unde is use, for subject ~~
% 환기
%
%Our method is selective because we have found that incorporating informative details of the \emph{subject} itself can have a detrimental impact, undermining the preservation of the subject's identity. 
Our method is \emph{selective} because we deliberately avoid incorporating informative specifications of the ``subject'' itself into the train descriptions. Such incorporation could negatively affect the preservation of the subject’s identity, as will be elucidated %illustrated
in \cref{sec:selectively informative descriptions}.
% Informative specifications should be included only for the undesired objects. 
%
SID can be readily integrated with any optimization-based model, such as DreamBooth~\cite{ruiz2023dreambooth}, Custom Diffusion~\cite{kumari2023multi}, or SVDiff~\cite{han2023svdiff}. For the automated generation of SID, we utilize the multimodal GPT-4~\cite{openai2023gpt4} by guiding it with an appropriate instruction.
Surprisingly, the proposed SID integration significantly reduces embedding entanglement, especially in critically biased scenarios.
\vspace{-0.1cm}

\vspace{-0.1cm}
\section{Related Work}
\vspace{-0.2cm}
\paragraph{Text-to-image diffusion models.} Diffusion models~\cite{song2019generative, ho2020denoising, song2020score} have demonstrated remarkable generative capabilities across a wide range of modalities, including images~\cite{dhariwal2021diffusion, nichol2021glide, saharia2022photorealistic}, audios~\cite{kong2020diffwave, chen2020wavegrad, liu2023audioldm}, videos~\cite{ho2022imagen, yang2022diffusion, luo2023videofusion}, and 3D shapes~\cite{zhou20213d, zeng2022lion, koo2023salad}. Within this diverse spectrum, prominent text-to-image diffusion models, such as Stable Diffusion~\cite{rombach2022high}, Imagen~\cite{saharia2022photorealistic}, and DALL-E 2~\cite{ramesh2022hierarchical}, have exhibited exceptional proficiency in the synthesis of high-quality images closely aligned with textual descriptions. Subsequently, significant efforts~\cite{chefer2023attend, feng2022training, phung2023grounded, liu2022compositional, lee2023aligning, black2023training} have been dedicated to enhancing the text-to-image alignment in the field of text-to-image generation. Our study, based on the Stable Diffusion model~\cite{rombach2022high}, improves the alignment between textual descriptions and images in personalized image synthesis, focusing on accurately preserving subject identity without any entanglement with undesired objects.

\vspace{-0.45cm}

\paragraph{Vision-language models.} 
The vision-language model (VLM) handles tasks like text-to-image retrieval~\cite{wang2019camp}, visual question answering~\cite{antol2015vqa}, and image captioning~\cite{xu2015show}. VLMs can also play a crucial role as a building block, as seen with CLIP~\cite{radford2021learning} in tasks such as text-to-image generation~\cite{ramesh2022hierarchical, nichol2021glide, rombach2022high}, image segmentation~\cite{luddecke2022image}, image retrieval~\cite{baldrati2022effective}, and more~\cite{patashnik2021styleclip, mohammad2022clip, shridhar2022cliport, singer2022make}. In our work, we need an image captioning VLM capable of closely following detailed instructions. While models like BLIP-2~\cite{li2023blip2} and OpenFlamingo~\cite{awadalla2023openflamingo} offer image captioning with text conditioning, they often struggle to adhere closely to detailed instructions. In contrast, large models such as LLaVA~\cite{liu2023visual} and multi-modal GPT-4~\cite{openai2023gpt4} have demonstrated their ability to generate precise captions closely aligned with detailed instructions. Among these options, multi-modal GPT-4 exhibits the most powerful capability in generating instruction-following captions. %leading us to select it as our final choice.

\vspace{-0.45cm}
\paragraph{Personalized image synthesis.}
Personalizing text-to-image diffusion models~\cite{chen2023subject, kumari2023multi, shi2023instantbooth, ruiz2023dreambooth, avrahami2023break, jia2023taming, wei2023elite, gal2022image, gal2023encoder, li2023blip, tewel2023key} has become a central focus in text-to-image generation. These models generate high-quality, diverse renditions of a subject using either a few reference images (usually 3 to 7) or just a single reference image. These models can be broadly categorized into two groups: optimization-based and encoder-based models. Optimization-based models~\cite{gal2022image, ruiz2023dreambooth, kumari2023multi, han2023svdiff, tewel2023key, avrahami2023break}, like DreamBooth~\cite{ruiz2023dreambooth}, involve per-subject optimization, where they fine-tune identifier tokens or segments of diffusion models to encode a subject's identity. On the other hand, encoder-based models~\cite{jia2023taming, wei2023elite, shi2023instantbooth, li2023blip, gal2023encoder} pre-train separate identity encoders and utilize them to encode a subject's identity without the need for per-subject optimization.

\newcommand{\quotes}[1]{``#1''}
\vspace{-0.15cm}
\section{Method}
\begin{figure}[t]
\centering
\includegraphics[width=0.8\hsize]{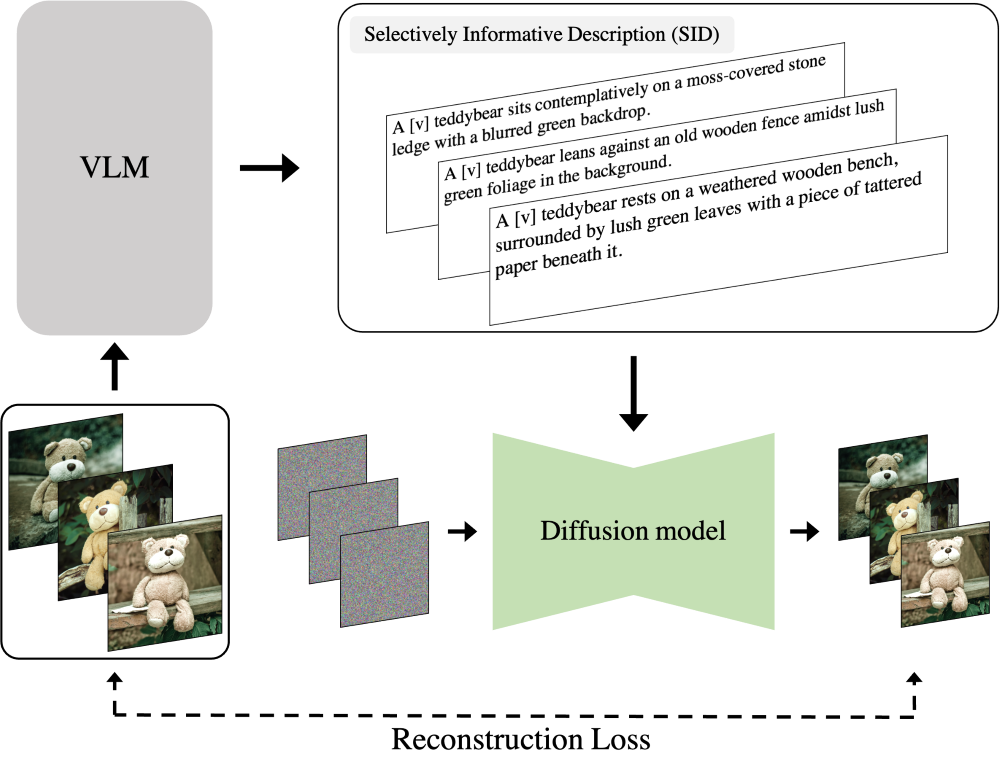}
\vspace{-0.3cm}
\caption{
\textbf{Personalization with SID}. We propose integrating SID~(Selectively Informative Description) into the per-subject optimization, where an instruction-following VLM~(Vison-Language Model) is utilized to generate a selectively informative description for each reference image. 
}
% \caption{
% \textbf{SID-integrated per-subject optimization}. We suggest integrating SID (Selectively Informative Description) into the per-subject optimization to reduce undesired embedding entanglement, especially when dealing with critically biased reference images. The reconstruction learning process utilizes SIDs generated by VLM (vison-language model), significantly enhancing subject and text alignment.
% }
\vspace{-0.6cm}
\label{fig:method}
\end{figure}

\vspace{-0.15cm}
Our proposed method can be readily integrated with any optimization-based models that utilize description of the reference images. The overall process, presented in \cref{fig:method}, entails the generation of \emph{SID (Selectively Informative Description)} for each reference image using an instruction-following vision-language model (VLM). These SIDs are integrated into the reconstruction learning process as part of the per-subject optimization. Depending on the underlying optimization-based model, additional loss functions may be incorporated together with the main reconstruction loss. The additional loss functions, such as the class prior preservation loss~\cite{ruiz2023dreambooth} introduced in DreamBooth, are applied in the same manner as in their respective baseline models without any modification. The effect of SID is addressed in \cref{sec:selectively informative descriptions}. The choice of VLM and its accompanying instruction are addressed in \cref{sec:integrating SID}.

\begin{figure*}[t]
\centering
\includegraphics[width=0.83\textwidth]{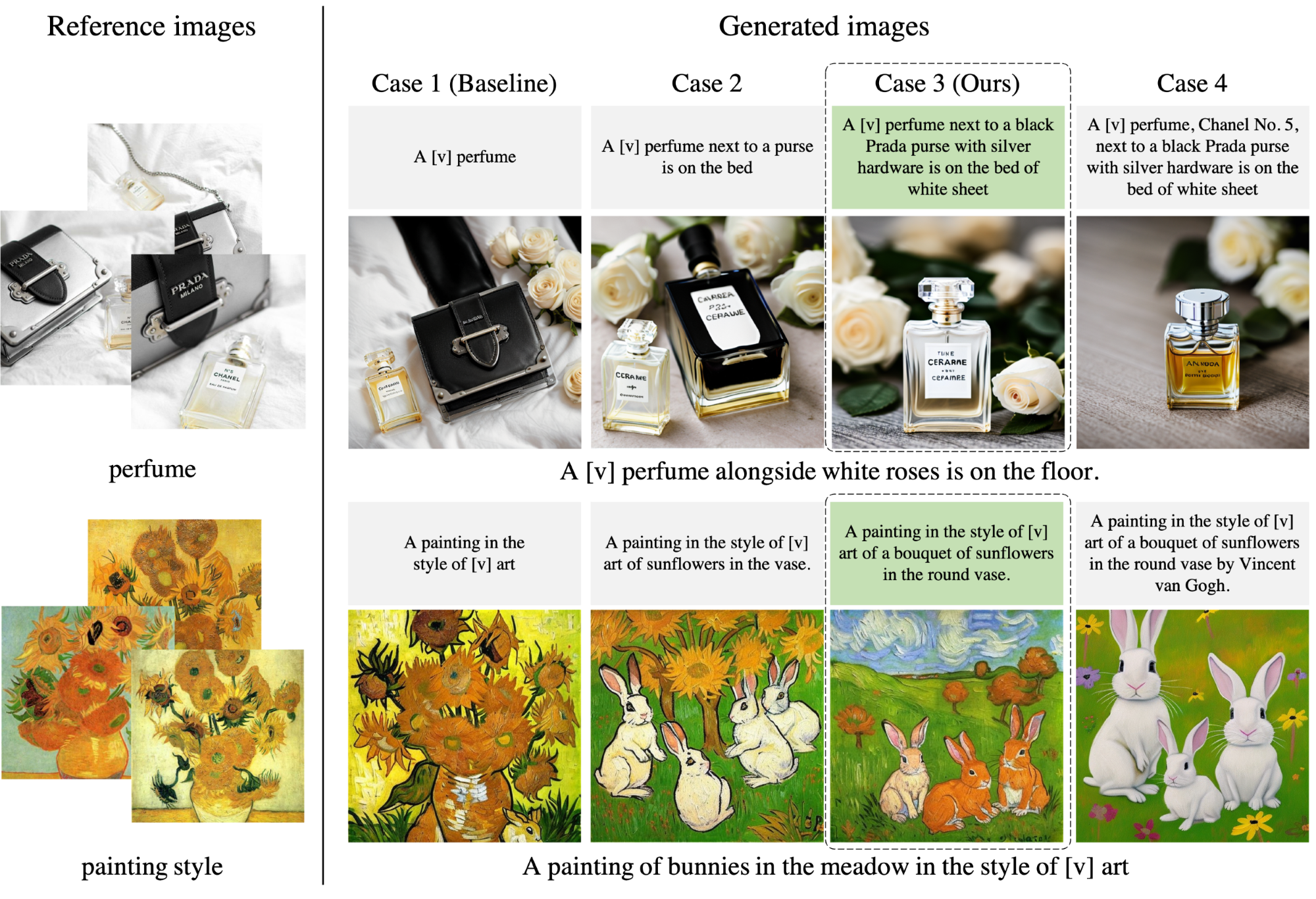}
\vspace{-0.1cm}
\caption{\textbf{Two examples for comparing the four cases of descriptions.} 
%\underline{Left}: Reference images. \underline{Right}: Generated images. 
For each case, the choice of train description follows the guidelines in \cref{tab:prompt_ablation}. The common generation prompt for each example is shown below the generated images. Additional examples can be found in \cref{sec: Supple_four description cases}.}
% \caption{\textbf{Qualtitative comparisons between the four descriptions cases.} \underline{Left}: Highly biased reference sets used for per-subject optimization, along with the subject’s coarse class presented below. \underline{Right}: Generated images for each description case, outlined in \cref{tab:prompt_ablation}, accompanied by the generation prompts beneath each row of images. Train descriptions are displayed above the generated images. In each description case, the same train descriptions are repeated across the reference images due to their substantial similarities. Case 3 (Ours) shows the most favorable subject and text alignments while accurately isolating the subjects. Case 4 fails to faithfully preserve the subject's identity. More examples can be found in the supplementary material.}
\label{fig:prompt_ablation}
\vspace{-0.4cm}
\end{figure*}

\subsection{SID for reducing embedding entanglement}
\vspace{-0.1cm}
\label{sec:selectively informative descriptions}
The existing text-to-image personalization approaches~\cite{ruiz2023dreambooth, gal2022image, kumari2023multi, han2023svdiff}, especially optimization-based methods like DreamBooth~\cite{ruiz2023dreambooth}, often rely on simplistic textual descriptions that closely follow the format of \quotes{a [v] [class name]}, where [v] denotes the unique identifier for the subject, in order to extract the subject's identity. In our method, our goal is to come up with train descriptions that have an appropriate level of information such that we can effectively isolate the subject of interest even when the reference images are biased.
According to the findings in DreamBooth~\cite{ruiz2023dreambooth}, specifying the subject's correct class in the train description is instrumental in accurately preserving the subject's identity. With the correct class information, the model's prior knowledge of the subject's class can significantly enhance the editing capabilities of the model. 
Based on the findings, we consider the case where the train description only contains the subject's class identification as the baseline and subsequently explore additionally informative cases that are listed in \cref{tab:prompt_ablation}. In \cref{tab:prompt_ablation}, the term \emph{undesired object} generally encompasses any object within reference images, excluding the subject of interest. 
For the examples in \cref{fig:main_bias}, the undesired objects can be readily recognized for background, nearby-object, and tied-object biases. For the substance bias, that is applicable only when performing a style re-contextualization, all the objects in the reference images are considered to be undesirable because the intention is to transfer the style only. For pose bias, it is inherent to the subject itself.

\begin{table}[t]
\resizebox{\columnwidth}{!}{%
\begin{tabular}{llcccc}
\toprule
\multicolumn{2}{l}{\multirow{2}{*}{Descriptions}} & \multicolumn{2}{c}{Subject}                                                         & \multicolumn{2}{c}{Undesired objects}                                               \\ \cmidrule(l){3-4} \cmidrule(l){5-6} 
\multicolumn{2}{c}{}                              & \begin{tabular}[c]{@{}c@{}}Class\\ Identification\end{tabular} & \begin{tabular}[c]{@{}c@{}}Informative\\ Specifications\end{tabular} & \begin{tabular}[c]{@{}c@{}}Class\\ Identification\end{tabular} & \begin{tabular}[c]{@{}c@{}}Informative\\ Specifications\end{tabular} \\ \midrule
\multicolumn{2}{l}{Case 1 (Baseline)}             & O              & X                                                                  & X              & X                                                                  \\
\multicolumn{2}{l}{Case 2}                        & O              & X                                                                  & O              & X                                                                  \\
\multicolumn{2}{l}{Case 3 (Ours)}                 & O              & X                                                                  & O              & O                                                                  \\
\multicolumn{2}{l}{Case 4}                        & O              & O                                                                  & O              & O                                                            
      \\ \bottomrule
\end{tabular}%
}
\vspace{-0.15cm}
\caption{\textbf{Four cases of train descriptions.} The four cases are categorized based on whether they provide class identification only or additional informative specifications of the subject and undesired objects. We define \textit{SID} as the description of Case 3 (Ours).}
\label{tab:prompt_ablation}
% \caption{\textbf{Outline of the four description cases.} The four cases are divided based on whether they include class identification or provide additional informative specifications of the subject or undesired objects. We define \textit{SID} as the description of Case 3 (Ours). Basically, any information not desired to be included in [v] should be described with the maximum amount of informative specifications, and any information desired to be included in [v] should \textit{not} be described in detail except for the basic class information.}
% \label{tab:prompt_ablation}
\vspace{-0.6cm}
\end{table}
For the four cases in~\cref{tab:prompt_ablation}, two examples are investigated in~\cref{fig:prompt_ablation} where two sets of the resulting images are shown. For the simplicity of this investigation, reference images were chosen to share the same set of undesired objects and thus allowing us to use a single train description for all reference images. Case 1 (Baseline) uses the format \quotes{a [v] [class name]}, where only the subject’s class identification is included in the train description. In this case, the model often struggles with the bias in the reference images (e.g., the subject can co-appear with undesired objects). This is because of the embedding entanglement in [v] as will be analyze further in \cref{sec:Analysis_cross}. 
\vspace{-0.05cm}
To reduce the undesired embedding entanglement, we introduce Case 2 where the train description contains the class identifications of the undesired objects. Case 3 is similar to Case 2, but the train description additionally contain informative specifications of the undesired objects.  
As shown in \cref{fig:prompt_ablation}, both Case 2 and Case 3 can provide a significant improvement over Case 1. Case 3, however, often outperforms Case 2 in terms of entanglement reduction as is the case for the two examples in \cref{fig:prompt_ablation}. 

For the subject itself, it is also possible to include informative specifications of the subject. Case 4 corresponds to this possibility. With the inclusion, however, the subject details described in the informative specifications are disentangled from the [v]. This occurs because of the alignment between the subject's informative specifications in the text and the subject details in the images. 
Because preserving the subject details in the [v] is crucial in personalized image synthesis, we have opted for Case 3 as our definition of SID.

\begin{figure}[t]
    \centering
\includegraphics[width=0.9\columnwidth]{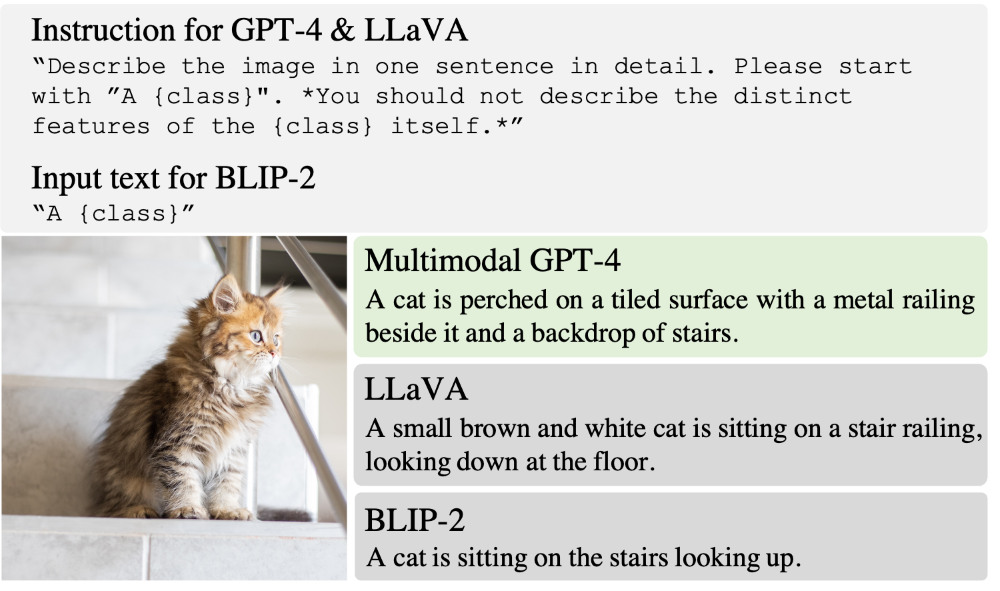}
% \vspace{-0.3cm}
    \vspace{-0.2cm}
    \caption{\textbf{Comparison of three instruction-following VLMs for generating SID.} For the reference image of a cat and the instructions shown in the top, the three VLMs generate image captions shown in the right side of the image. Subsequently, the unique identifier [v] is inserted and the resulting captions are used for conditioning the diffusion model as the train descriptions. 
    For painting/cartoon style re-contextualization, we used a slightly different instruction, as detailed in \cref{sec: Supple_for style re-contextualization}.
    Additional fifteen examples for VLM-generated SIDs are available in \cref{sec: Supple_instruction-following VLMs}.}
    \label{fig:VLM_example}
    % \caption{\textbf{Comparison of instruction-following VLMs for image captioning.} Evaluating BLIP-2~\cite{li2023blip2}, LLaVA~\cite{liu2023visual}, and multi-modal GPT-4~\cite{openai2023gpt4}, the GPT-4 demonstrates the closest alignment with the concept of SIDs. Captions (right) are generated based on provided instructions (top) and an input image (left). More examples are available in the supplementary material.}
    % \label{fig:VLM_example}
\vspace{-0.5cm}
\end{figure}

\subsection{VLM for generating SID}
% \vspace{-0.2cm}
\label{sec:integrating SID}
\vspace{-0.1cm}
Generating SID for each reference image can be demanding and time-consuming for humans. Therefore, we have chosen to utilize image captioning VLM for an automatic generation of SID. 
To identify a suitable VLM capable of effectively generating SIDs, we evaluated three well-known image captioning VLMs: BLIP-2~\cite{li2023blip2}, LLaVA~\cite{liu2023visual}, and multi-modal GPT-4~\cite{openai2023gpt4}. In \cref{fig:VLM_example}, we illustrate a caption generation example in which the subject is explicitly defined as a cat within the image. For BLIP-2, it struggles to comprehend detailed instructions and it can often produce sequences of meaningless characters. Therefore, we exclusively conditioned BLIP-2 for the subject of interest while providing the detailed instructions to both LLaVA and GPT-4 for generating SID. 
% \vspace{-0.05cm}

Captions generated by BLIP-2 significantly fall short of the desired SID, lacking informative specifications, i.e., detailed features, of any undesired objects (objects other than the cat in \cref{fig:VLM_example} example) in the images. Although LLaVA generates descriptions better aligned with the instructions, it still falls substantially short, tending to describe the subject excessively or fail to provide sufficient details about undesired objects. Compared to the other two, multi-modal GPT-4 stands out for its ability to follow instructions, delivering descriptions that closely align with the concept of SID. The quantitative comparisons between these VLMs are shown in \cref{sec: Supple_instruction-following VLMs}, where multi-modal GPT-4 shows the best subject and text alignment. Therefore, we have chosen multi-modal GPT-4 as the VLM for SID generation.

\vspace{-0.2cm}
\section{Experiments}
\label{sec:Experiments}
\vspace{-0.2cm}
% In this section, we illustrate how integrating SID into per-subject optimization effectively reduces undesired embedding entanglement, with a particular focus on reference sets exhibiting high biases. We present qualitative analysis in \cref{sec:qualitative analysis} and quantitative analysis in \cref{sec:quantitative analysis}.

\begin{figure*}[t]
    \centering
    \includegraphics[width=0.8\textwidth]{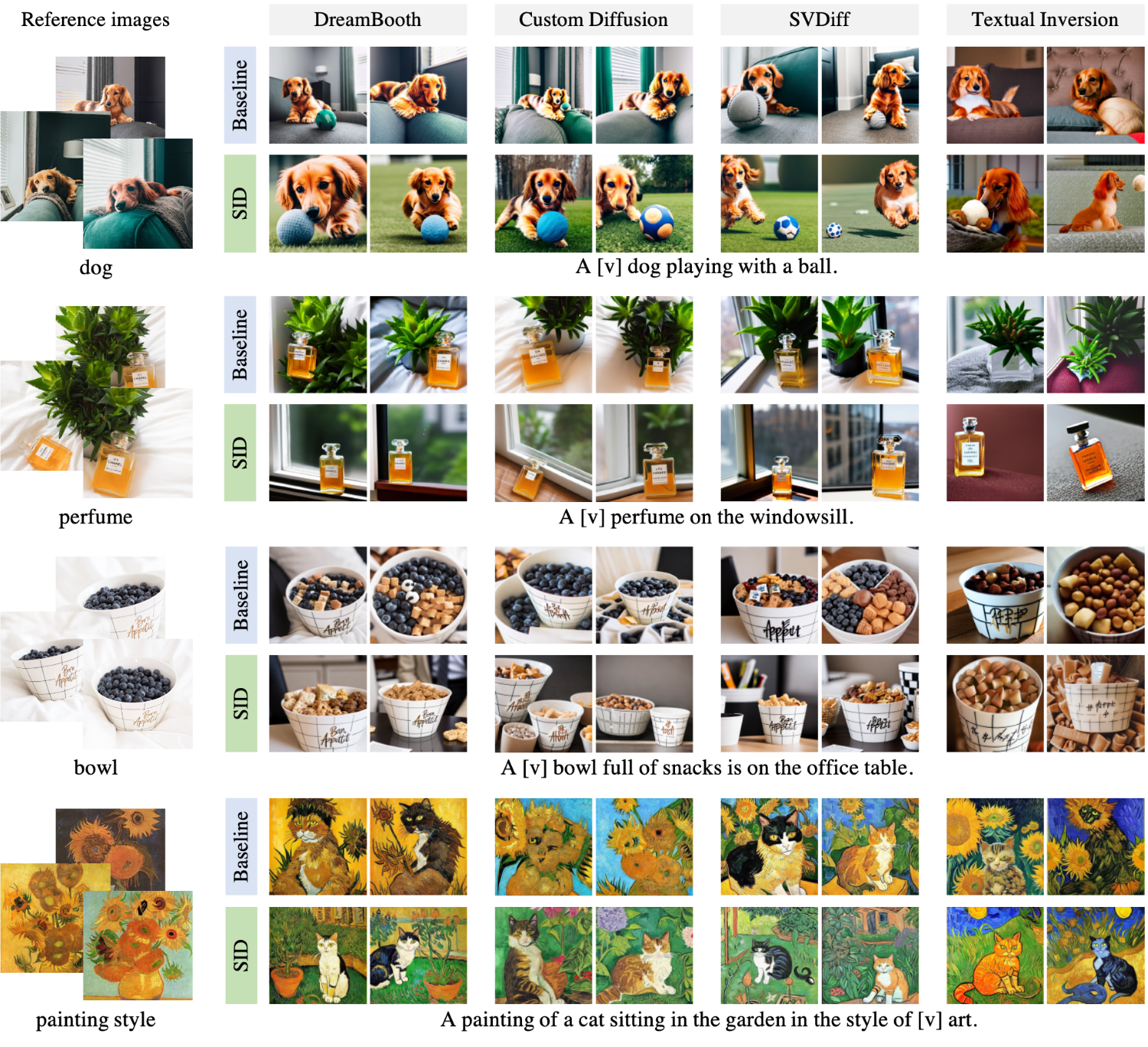}
    % 0.852
    \vspace{-0.3cm}
    \caption{
    \textbf{Enhancement by SID.} For four optimization-based models (DreamBooth~\cite{ruiz2023dreambooth}, Custom Diffusion~\cite{kumari2023multi}, SVDiff~\cite{han2023svdiff}, and Textual Inversion~\cite{gal2022image}), the baseline results are shown together with SID-integrated results. SID-integration effectively resolves entanglement issues in scenarios with high biases, represented by indoor background (1st row), nearby potted plant (2nd row), filled-in blueberries (3rd row), and sunflower substances (last row). Additional examples can be found in \cref{sec: Supple_enhancement by SID}.
    }
    \vspace{-0.6cm}
    % \caption{
    % \textbf{Qualitative comparison} of optimization-based models (DreamBooth~\cite{ruiz2023dreambooth}, Custom Diffusion~\cite{kumari2023multi}, SVDiff~\cite{han2023svdiff}, and Textual Inversion~\cite{gal2022image}) and their SID-integrated counterparts in biased scenarios. Each row displays reference images with a coarse class name (left) and corresponding generated images from the optimization-based models and their SID-integrated counterparts (right). A single generation prompt is utilized for each row. SID-integration effectively addresses entanglement issues in diverse scenarios, such as indoor backgrounds (1st row), nearby potted plants (2nd row), filled-in blueberries (3rd row), and sunflower substance (last row), while demonstrating superior text alignment compared to baselines. More examples can be found in the supplementary material.
    % }
    \label{fig:vs_optimization-based}
\end{figure*}

\begin{figure}[h]
    \centering
\includegraphics[width=0.85\columnwidth]{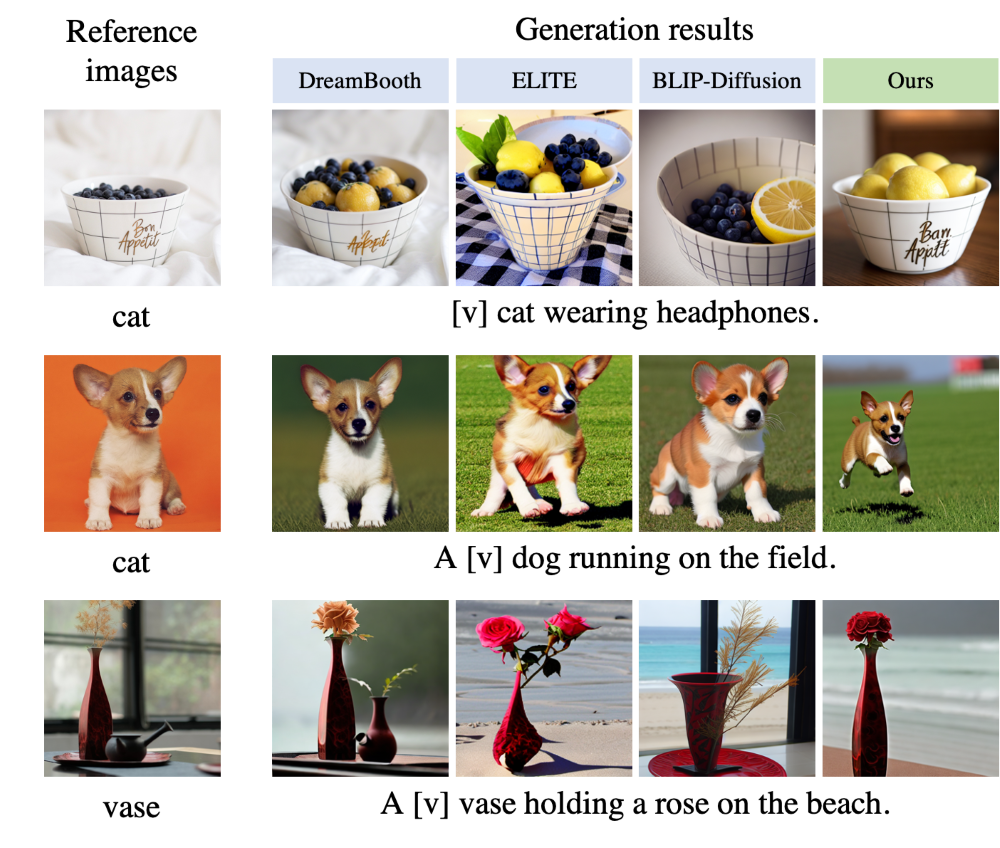}
\vspace{-0.4cm}
    \caption{\textbf{Enhancement by SID for a single reference image.}  Two encoder-based models, ELITE~\cite{wei2023elite} and BLIP-Diffusion~\cite{li2023blip}, are compared with DreamBooth~\cite{ruiz2023dreambooth} and our SID-integrated DreamBooth. Notably, in the second row, our approach effectively addresses pose bias. Additional examples can be found in \cref{sec: Supple_enhancement for a single reference image}.
    }
    % \caption{\textbf{Comparison on single reference images.} Compared to DreamBooth~\cite{ruiz2023dreambooth}, or encoder-based models like ELITE~\cite{wei2023elite} and BLIP-Diffusion~\cite{li2023blip}, SID-integrated DreamBooth (Ours) accurately preserves the subject’s identity while significantly reducing undesirable entanglements, resulting in \emph{Pareto improvement}.}    
    \vspace{-1cm}
    \label{fig:single reference images}
\end{figure}

We have performed comprehensive experiments to verify the enhancement resulting from SID. For the experiment datasets, images from previous works~\cite{ruiz2023dreambooth, kumari2023multi, gal2022image} and three websites~\cite{Unsplash, peppercarrot, wikiart} were examined. As the main experiment, we have integrated SID with four optimization-based models -- DreamBooth~\cite{ruiz2023dreambooth}, Custom Diffusion~\cite{kumari2023multi}, SVDiff~\cite{han2023svdiff}, and Textual Inversion (TI)~\cite{gal2022image}. %All models are based on the Stable Diffusion architecture \cite{rombach2022high}.
% The results are shown in \cref{fig:vs_optimization-based}, with additional results in \textcolor{orange}{Supplementary ABC}. 
The superior effectiveness of SID is readily apparent across a broad spectrum of examples as shown in \cref{fig:vs_optimization-based}. 
In the presence of a single reference image only, we have also compared SID-integrated DreamBooth (Ours) with two encoder-based models, ELITE~\cite{wei2023elite} and BLIP-Diffusion~\cite{li2023blip}, along with DreamBooth in \cref{fig:single reference images}.
Implementation details for each model are provided in \cref{sec: Supple_implementation details}.

% The results are shown in \cref{fig:single reference images}, with additional results in \textcolor{orange}{Supplementary ABC}. 

%As in \cref{fig:vs_optimization-based}, the superiority of SID-integrated model can be readily confirmed. 
%DreamBooth exhibits strong subject-alignment, but faces challenges disentangling undesirable objects in the reference images. Encoder-based models slightly reduce these entanglements but also compromise the subject's identity. In contrast, SID-integrated DreamBooth (Ours) accurately preserves the subject’s identity while significantly reducing undesirable entanglements, leading to a \emph{Pareto improvement}, where high subject and text alignments are simultaneously achieved.

\vspace{-0.2cm}
\section{Analysis of cross-attention map}
\label{sec:Analysis_cross}
\begin{figure}[t]
    \centering
    \includegraphics[width=0.95\columnwidth]{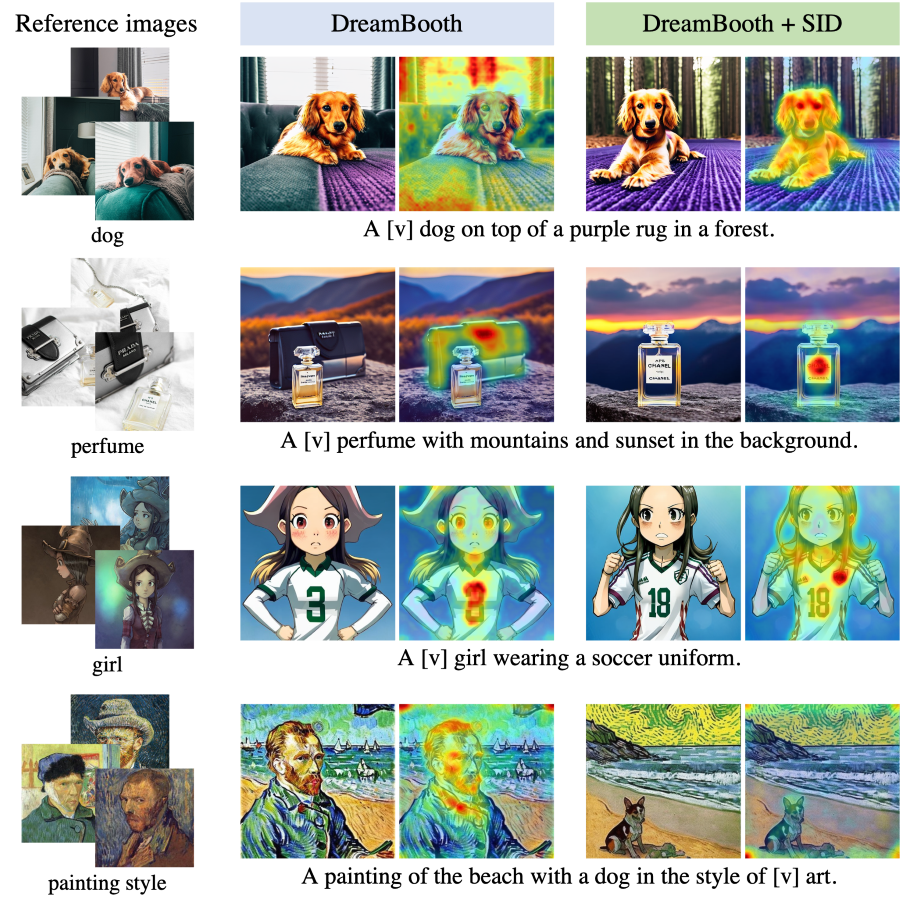}
    \vspace{-0.2cm}
    \caption{\textbf{Cross-attention map.} Averaged cross-attention maps for the identifier [v] are shown. They are overlayed on the generated images. 
    The undesirably entangled embeddings of DreamBooth~\cite{ruiz2023dreambooth} can be confirmed from the erroneous focusing on the undesirable objects. 
    On the contrary, SID-integrated DreamBooth accurately highlights the subjects of interest: dog, perfume, girl, and painting style. 
    % \jimyeong{We also visualized cross-attention map analysis for the four cases described in \cref{fig:prompt_ablation}.}
    Image credit: \href{https://www.peppercarrot.com/}{David Revoy} (3rd row)}
    % \caption{\textbf{Reducing undesired embedding entanglements.} Averaged cross-attention maps for the identifier [v] are overlayed on generated images. DreamBooth~\cite{ruiz2023dreambooth} shows undesirably entangled embeddings, erroneously focusing on undesirable objects, while SID-integrated DreamBooth accurately highlights the subjects of interest: dog, perfume, girl, and painting style.}
    \label{fig:heatmap}
    \vspace{-0.5cm}
\end{figure}

\vspace{-0.2cm}
The cross-attention map of each text token is known to highlight the pixels the text token describes~\cite{hertz2022prompt, tang2022daam}. We visualized averaged cross-attention maps for the identifier [v], which is trained to encode the subject’s identity. The cross-attention maps are averaged over all the timestamps, layers, and heads to show the single image that can be overlayed on the generated image, following the previous convention~\cite{tang2022daam}. In \cref{fig:heatmap}, we compared cross-attention maps for DreamBooth and SID-integrated DreamBooth across four distinct biases: background, nearby-object, tied-object, and substance biases. While DreamBooth’s identifier tends to erroneously focus on undesired objects across all biased scenarios, our method consistently displays highly accurate attention focused on the subject. To extend this analysis, we conducted a comparison between the four cases of train descriptions (detailed in \cref{fig:prompt_ablation}, top row) in \cref{fig:rebuttal_attention}. These analyses confirm that integrating SID does reduce undesired embedding entanglements.

\vspace{-0.1cm}
\section{Analysis of three key measures}
\label{sec:Analysis_measures}
\vspace{-0.1cm}
Image-alignment and text-alignment have become popular measures in recent personalized image synthesis~\cite{chen2023subject, kumari2023multi, shi2023instantbooth, ruiz2023dreambooth, avrahami2023break, jia2023taming, wei2023elite, gal2022image, gal2023encoder, li2023blip, tewel2023key}. However, we have found that the widely used image-alignment score tends to be significantly influenced by background entanglement, making it an inappropriate measure for analyzing undesired embedding entanglements. Therefore, we first define two customized measures, \emph{subject-alignment} and \emph{non-subject-disentanglement}, by adapting image-alignment specifically for text-to-image personalization. Further details can be found in \cref{sec: Supple_evaluation metrics}, where we demonstrate the crucial importance of these two measures in assessing subject preservation and reduction in undesired embedding entanglement. 
Using the two measures, along with text-alignment, we conduct a quantitative analysis of SID.
\vspace{-0.2cm}

\paragraph{Customizing image-alignment measure.} Consider a set of reference images $R = \{r_1, r_2, \cdots, r_N\}$ and a set of generated images $G = \{g_1, g_2, \cdots, g_M\}$. Each image in $G$ is generated using $R$ as the reference images with a common generation prompt $p$. Define $f_i: \mathcal{X}_{image} \rightarrow \mathbb{R}^D$ and $f_t: \mathcal{X}_{text} \rightarrow \mathbb{R}^D$ as the CLIP image and text encoders~\cite{radford2021learning} followed by a unit-norm normalization, respectively. Also, define a subject segmentation function $s: \mathcal{X}_{image} \rightarrow \{0, 1\}^{H \times W}$, implemented as Grounded-SAM~\cite{Grounded-SAM_Contributors_Grounded-Segment-Anything_2023} conditioned on the subject's class name. A mask over the image resolution space is output by $s$ where 1 indicates the subject's pixel and 0 indicates the non-subject's pixel.

% The subject-alignment $\mathcal{SA}$ is defined as the averaged pairwise cosine similarity of subject segments between reference and generated images in the CLIP-embedding space.
The subject-alignment $\mathcal{SA}$ is defined as the average pairwise cosine similarity between generated images and subject segments of reference images in the CLIP-embedding space.
The non-subject-disentanglement $\mathcal{NSD}$ is defined as the 1 minus the non-subject segment similarity in CLIP-embedding space. They can be formally expressed as:
%
% \begin{align}
%     \mathcal{SA} =& Avg ( f_i(r_n \odot s (r_n))^\intercal \cdot f_i(g_m \odot s (g_m)) ) \\
%     \mathcal{NSD} =& 1 - Avg( f_i(r_n \odot (1 - s (r_n)))^\intercal \nonumber \\
%     &  \ \ \ \ \ \ \ \ \ \ \ \ \ \ \ \cdot f_i(g_m \odot (1 - s (g_m))) ),
% \end{align}
\begin{align}
    \mathcal{SA} &= \ \ \ \ \ \ \ Avg( f_i(r_n \odot s (r_n))^\intercal \cdot f_i(g_m) ), \\
    \mathcal{NSD} &=1-Avg( f_i(r_n \odot (1 - s (r_n)))^\intercal  \cdot  f_i(g_m) ),
\end{align}
where $Avg(\cdot)$ denotes the average operation over all possible pair selections between $n=1,\dots,N$ and $m=1, \dots, M$. Additionally, we evaluate text-alignment $\mathcal{TA}$ shown below. 
\begin{equation}
    \mathcal{TA} =  Avg ( f_t(p)^\text{T} \cdot f_i(g_m) ).
\end{equation}
% \vspace{-0.1cm}
The identifier [v] is removed from the generation prompt $p$ when evaluating $\mathcal{TA}$.
\vspace{-0.2cm}

%\begin{equation}
%    \text{SA} = \frac{1}{N \cdot M} \sum_{n = 1}^{N} \sum_{m = 1}^{M} \text{cos-sim}( f_i(r_n \odot f_s (r_n)), f_i(g_m \odot f_s (g_m)) )
%\end{equation}
% \begin{equation}
%     \text{NSD} = 1 - \frac{1}{N \cdot M} \sum_{n = 1}^{N} \sum_{m = 1}^{M} \text{cos-sim}( f_i(r_n \odot (1 - f_s (r_n))),\newline f_i(g_m \odot (1 - f_s (g_m))) )
% \end{equation}
%\begin{equation}
%    \text{TA} = \frac{1}{N \cdot M} \sum_{n = 1}^{N} \sum_{m = 1}^{M} \text{cos-sim}( f_t(p), f_i(g_m) )
%\end{equation}

\begin{figure}[t]
\centering
\includegraphics[width=0.95\columnwidth]{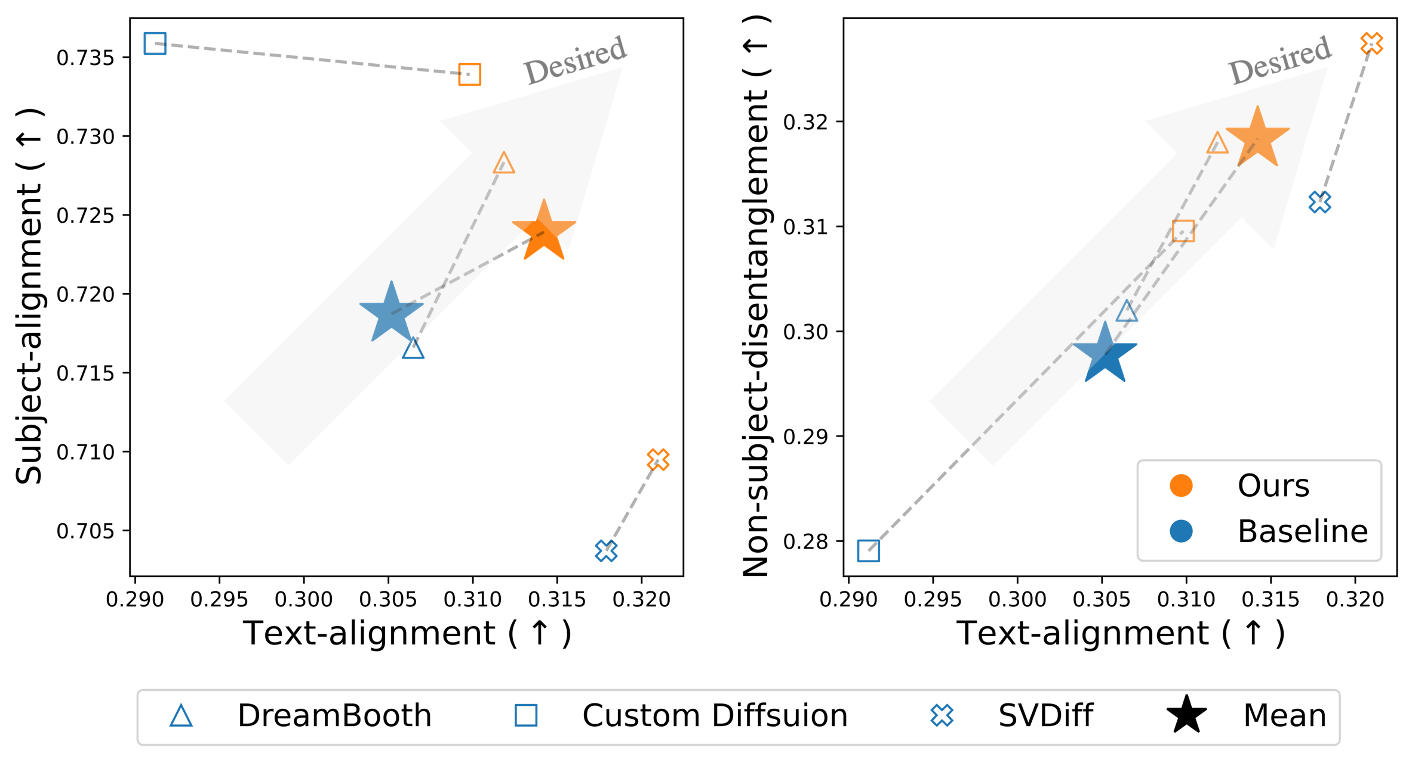}
\vspace{-0.2cm}
\caption{\textbf{Pair-wise metric visualization for multiple reference images.} The best performing ones are the SID-integrated models and they form the Pareto boundary.}
% \caption{\textbf{Quantitative comparison in multiple reference images.} SID-integration across all four optimization-based models enhances personalized image synthesis, effectively reducing undesired entanglements (increased \emph{non-subject-disentanglement}).}
\label{fig:multiple_reference_curve}
\end{figure}

\begin{figure}[t]
\centering
\includegraphics[width=0.95\columnwidth]{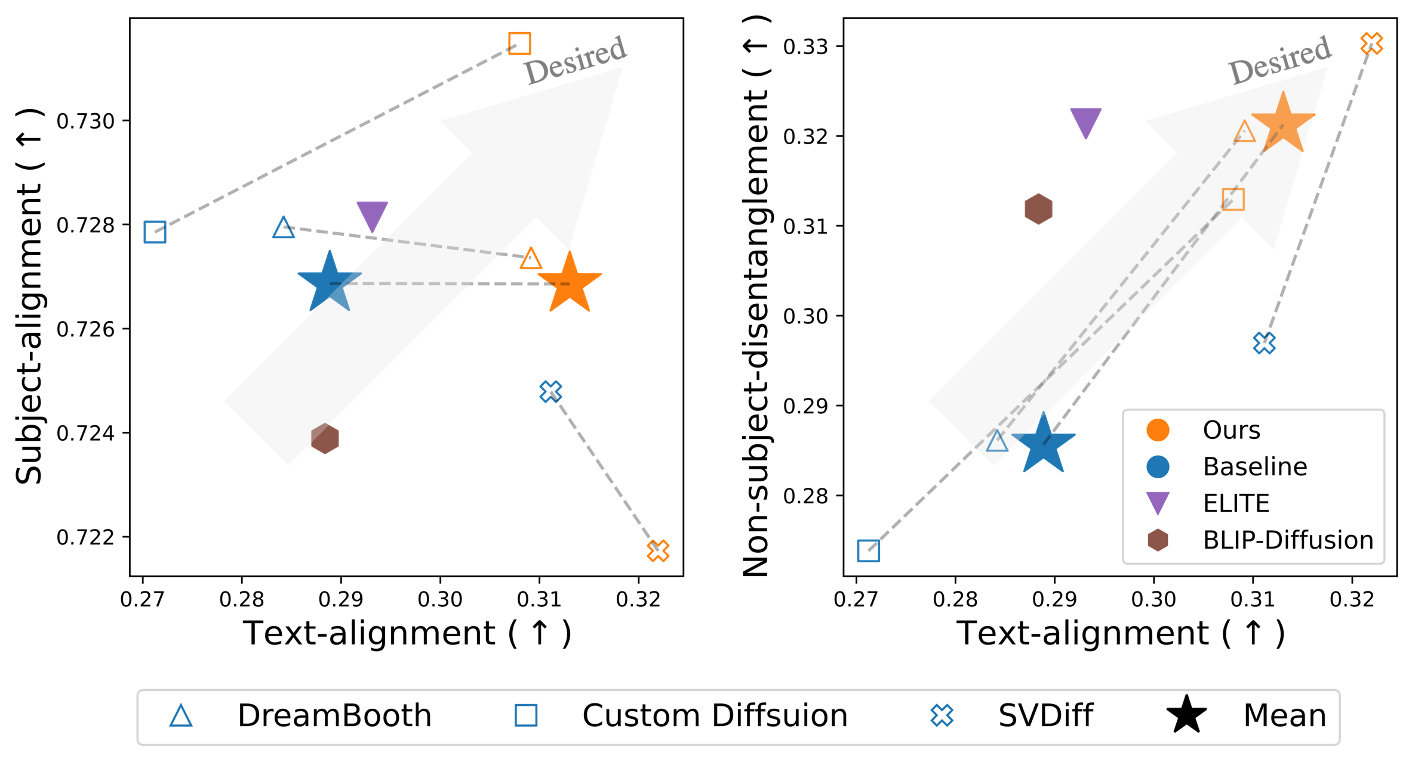}
\vspace{-0.2cm}
\caption{\textbf{Pair-wise metric visualization for single reference image.} The best performing ones are the SID-integrated optimization-based models and they form the Pareto boundary. They also outperform the two encoder-based models.}
\vspace{-0.5cm}
% \caption{\textbf{Quantitative comparison in single reference images.} SID-integration enhances personalized image synthesis in DreamBooth~\cite{ruiz2023dreambooth}, Custom Diffusion~\cite{kumari2023multi}, and SVDiff~\cite{han2023svdiff}, outperforming ELITE~\cite{wei2023elite} and BLIP-Diffusion~\cite{li2023blip}. All four SID-integrated models exhibit reduced undesired entanglements (right).}
\label{fig:single_reference_curve}
\end{figure}

\paragraph{Analysis of SID.} The three measures have straightforward interpretations. A higher subject-alignment indicates superior preservation of the subject's identity. Elevated non-subject-disentanglement values signify a reduction in undesired embedding entanglements. Increased text-alignment suggests that the generated images align more closely with the generation prompt.
% using the three key measures and the results are
Using the three measures, we have evaluated the three optimization-based models and their SID-integrated counterparts as shown in~\cref{fig:multiple_reference_curve}. For the single reference images, we evaluated ELITE~\cite{wei2023elite} and BLIP-Diffusion~\cite{li2023blip} together with the three optimization-based models and the results are shown in~\cref{fig:single_reference_curve}. 
From the figures, it can be observed that all of the best performing, or \emph{Pareto optimal}, models are the SID-integrated models. The average improvements of the three key metrics exhibit positive values with a substantial margin in all cases, except for subject alignment in~\cref{fig:single_reference_curve}. 
A potential explanation for this observation is the occurrence of subject overfitting in the existing models, particularly evident in cases such as pose bias. The implementation of SID mitigates subject overfitting, even when only a single reference image is provided.
For the evaluation, we have utilized commonly used personalization datasets~\cite{kumari2023multi,ruiz2023dreambooth,gal2022image} to generate $7500$ images. The details are articulated in \cref{sec: Supple_datasets}.
%This is because of the Textual Inversion~\cite{gal2022image} that has low metric values even before SID is integrated. 

\vspace{-0.4cm}
\paragraph{Human evaluation.}
We further assess the effectiveness of SID over its baseline by conducting human evaluation.
Human evaluation was performed involving 130 participants, 10 subjects per participant, and 4 questions per subject (a total of 5,200 responses).
\cref{tab: human evaluation} shows that the survey results for subject-alignment, non-subject disentanglement, and text-alignment are consistent with the metric analysis in \cref{fig:multiple_reference_curve}, supporting our metric analysis. 
The results and additional details are provided in \cref{sec: Supple_human_evaluation}

\begin{figure*}[t]
    \centering
    \includegraphics[width=0.95\textwidth]{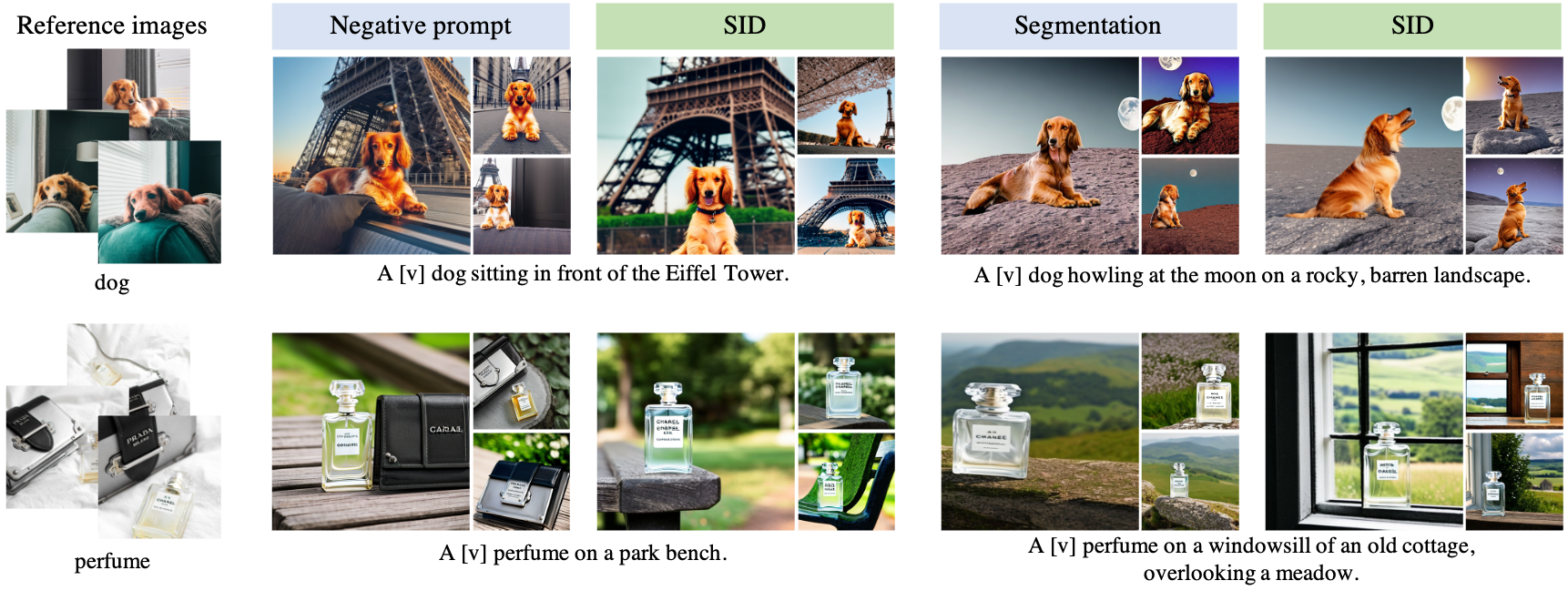}
    \vspace{-0.3cm}
    \caption{\textbf{Comparison with negative prompt and segmentation.} DreamBooth~\cite{ruiz2023dreambooth} is used as the base model. Compared to the two alternatives, SID-integration demonstrates superior non-subject-disentanglement and text-alignment. Negative prompts used: ``sitting on the fluffy blanket and green couch’' in the upper row example, ``next to a black Prada purse with silver hardware on a bed of white sheets'' in the lower row example.
    %\textcolor{blue}{For the negative prompt, we employed ``sitting on the fluffy blanket and green couch" for the dog and ``next to a black Prada purse with silver hardware on a bed of white sheets" for the perfume.} 
    Additional examples can be found in \cref{sec: Supple_negative prompt and segmentation}}
    % \caption{\textbf{Comparison with simple alternatives.} When compared to using negative prompts during inference or segmentation masks in per-subject optimization, SID demonstrates superior subject and text alignments. All the techniques are applied to DreamBooth~\cite{ruiz2023dreambooth}, and additional examples can be found in the supplementary material.}    
    \label{fig:vs_Neg_Seg}
    \vspace{-0.5cm}
\end{figure*}
\begin{figure}[t]
    \centering
    \includegraphics[width=\columnwidth]{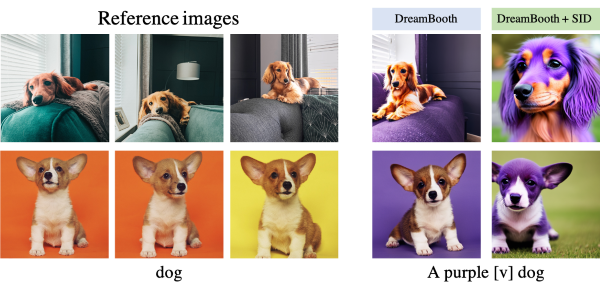}
    \vspace{-0.5cm}
    \caption{\textbf{Subject editing.} Two examples are shown for editing subject's color.
    DreamBooth~\cite{ruiz2023dreambooth} erroneously changes the color of non-subject elements (sofa in the upper row example, background in the lower row example), while SID-integrated DreamBooth accurately edits the subject (dog).}
    % \caption{\textbf{Undesired entanglements in color editing.} When subject embedding is undesirably entangled, misguided editing can happen. DreamBooth~\cite{ruiz2023dreambooth} erroneously changes the color of non-subject elements (sofa in the top row, background in the bottom row), while the SID-integrated counterpart accurately edits the subject (dog).}    
    \label{fig:discussion_color}
    \vspace{-0.4cm}
\end{figure}

\section{Discussion}
\label{sec:discussions}
%\paragraph{Comparison with simple alternatives.} 
\subsection{Negative prompt and segmentation} 
\label{sec:sub_sec:negativePrompt_seg}

Negative prompt for classifier-free guidance~\cite{ho2022classifier} and segmentation can be considered as the alternatives of SID. Negative prompt is typically used to prevent the generation of unwanted features~\cite{tumanyan2023plug, schramowski2023safe, wu2023human}. Segmentation mask has been used in several previous works for the purpose of isolating subject~\cite{jia2023taming, avrahami2023break, wei2023elite, shi2023instantbooth, li2023blip}.
Instead of adopting SID, negative prompt can be used during inference or the subject's segmentation mask can be applied before per-subject optimization.  
In \cref{fig:vs_Neg_Seg}, SID-integration is compared to these two approaches. Negative prompts exhibit unsatisfactory disentanglement, suggesting that severely entangled representations cannot be easily unraveled during inference. Segmentation masks also exhibit its own limitations, 
and face challenges in dynamically editing the pose of the subject or faithfully following the generation prompt. SID shows higher non-subject-disentanglement and text-alignment than the two alternatives.

\vspace{-0.1cm}
\subsection{Enhancing subject editing}
% \subsection{Undesired Entanglements in Subject Editing}
\vspace{-0.2cm}
The identifier [v], specifically trained to encode the subject's identity, enables us to modify the properties of the subject through a textual guidance. 
Undesired embedding entanglements, however, can lead to misaligned modifications. 
In \cref{fig:discussion_color}, an example of subject's color editing is shown. DreamBooth~\cite{ruiz2023dreambooth}, influenced by embedding entanglements, fails to edit the subject and instead modifies the color of an undesired object or background. In contrast, SID-integrated DreamBooth precisely modifies the subject itself, following the provided color guidance. %This underscores the potential threat associated with handling subject identity using undesirably entangled embeddings, emphasizing the critical importance of reducing undesired embedding entanglements.

\vspace{-0.1cm}
\subsection{Limitations of SID}
% \subsection{Failure Cases}
\label{sec:failure cases}
\vspace{-0.1cm}
A limitation of our approach is associated with the imperfections of VLM. While the multi-modal GPT-4~\cite{openai2023gpt4} generally generates descriptions that closely align with the provided instructions, occasional failures do occur. Although these instances are rare and may not be easily generalized, we have listed some of these failure cases in \cref{sec: Limitation_GPT_4_failure}. 
Another limitation is the undesired entanglement of the subject's strong facial expressions.
We discovered that this failure arises when the VLM-generated SID lacks informative specifications of facial expressions. This issue can be addressed by including undesired information in the SID, a solution detailed further in \cref{sec: Limitation_facial_expression}.
% Orig:
% In the last row of \cref{fig:main_bias}, SID-integrated DreamBooth successfully disentangles the dog's facial expression for a single reference image scenario. In the example, the dog in the reference image does not have any strong facial expression. When the reference images have a strong facial expression, however, SID-integrated models tend to struggle as shown in \cref{fig:facial_expression}.
Finally, SID is exclusively integrated with optimization-based models in this study. 
% In this study, SID is exclusively integrated with optimization-based models.
Yet, there is potential for its integration into encoder-based models, particularly during encoder pre-training.
% However, there exists potential for its incorporation into encoder-based models, particularly during the encoder pre-training. 
%In our upcoming research, we aim to explore this direction, expanding the versatility and scope of our approach.

\vspace{-0.20cm}
\section{Conclusion}
\vspace{-0.15cm}
In this study, we have introduced a robust strategy to mitigate undesired embedding entanglement in text-to-image personalization. Beginning with the identification of five key biases, we have proposed SID (Selectively Informative Description) as an effective solution to address these biases. Our cross-attention analysis demonstrates the successful removal of entanglements, while alignment analysis indicates notable enhancements in non-subject-disentanglement and text-alignment. The proposed SID strategy holds potential applicability to other multi-modal applications where managing embedding entanglements is crucial.

\vspace{-0.5cm}
\paragraph{Acknowledgement} This work was supported by the following grants funded by the Korea government: NRF (NRF-2020R1A2C2007139, NRF-2022R1A6A1A03063039) and IITP ([NO.2021-0-01343, Artificial Intelligence Graduate School Program (Seoul National University)], [No. RS-2023-00235293]).

{\small
%\bibliographystyle{ieeetr}
%\bibliographystyle{unsrt}
% \bibliographystyle{ieee_fullname}

% \bibliography{egbib}
% \bibliographystylelatex{ieee_fullname}
% \bibliographylatex{egbib}
}
% \clearpage
% \setcounter{page}{1}
% \setcounter{figure}{0}
% \setcounter{table}{0}
\appendix

\maketitlesupplementary

\renewcommand{\thetable}{\thesection.\arabic{table}}
\renewcommand{\thefigure}{\thesection.\arabic{figure}}
% \begin{append}
%\input{Suppl_TEX/00.Quantitative_analysis_Fig2}
%\input{Suppl_TEX/10.VLM_examples}

% \begin{refsection}
\setcounter{figure}{0}
\section{Evaluation measures}
\label{sec: Supple_evaluation metrics}
\begin{figure*}[t]
    \centering
    \includegraphics[width=0.85\textwidth]{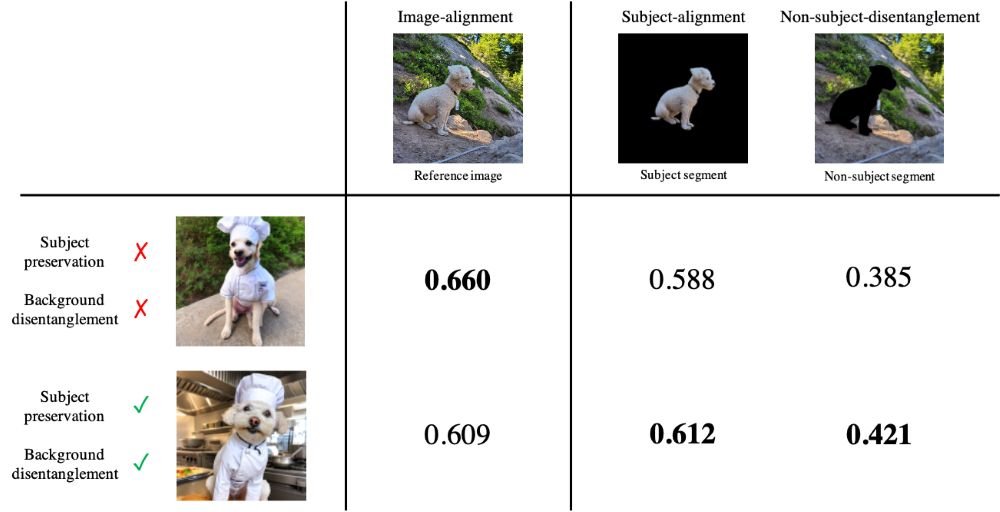}
    \caption{\textbf{A comparison between image-alignment and two customized measures.} The top row shows the reference image and its two segmented versions, while the left column shows two images generated with the prompt ``A [v] dog in a chef outfit.'' Despite poor subject preservation, the upper image in the left column obtains a higher image-alignment score because of the background bias. In contrast, subject-alignment effectively assesses subject preservation and non-subject-disentanglement accurately identifies background disentanglement.}
    \label{fig:comparing metrics with a dog example}
    \vspace{-0.5cm}
\end{figure*}
% \subsection{Subject-alignment \& Non-subject disentanglement}

%Our research aims to reduce undesired embedding entanglements resulting from non-subject biases in reference images. 
To evaluate embedding entanglements, we introduced two measures: \emph{subject-alignment} and \emph{non-subject-disentanglement}. These measures are obtained from the widely used image-alignment score, which calculates the cosine similarity between reference and generated images in the CLIP~\cite{radford2021learning} (or DINO~\cite{caron2021emerging}) embedding space. Unlike the image-alignment score, subject-alignment and non-subject-disentanglement utilize segmented reference images to assess similarities associated with subject segments or non-subject segments, providing clear insights into both subject preservation and the presence of undesired entanglements. 

In \cref{fig:comparing metrics with a dog example}, we compare image-alignment measure with our two customized measures using a dog example. The second and third rows in the figure display scores for two images generated by a personalized text-to-image diffusion model. The first row shows an image that fails to preserve the subject identity, with an undesirable background entanglement (natural landscape). The second row shows an image that successfully preserves the subject identity while disentangling the undesirable background. Our example in \cref{fig:comparing metrics with a dog example} reveal that the image-alignment can be significantly influenced by background information, raising concerns about its accuracy when measuring subject preservation in highly biased scenarios. In contrast, subject-alignment accurately evaluates subject preservation and non-subject-disentanglement effectively identifies background entanglement in \cref{fig:comparing metrics with a dog example}. These differences make the two customized measures well-suited for analyzing text-to-image personalization models in any level of biased scenarios.

While the two customized measures effectively capture subject preservation and undesired entanglements, they have limitations in evaluating style re-contextualization. Specifically, the two measures rely on segmentation in the image pixel space, making it challenging to separate style from the image itself. Moreover, quantifying style similarities while disregarding the content presented in the image is not a straightforward task. Consequently, in all the quantitative analyses presented in this paper, we have excluded scenarios involving painting/cartoon style re-contextualization, even though our approach demonstrates significant visual improvements. As part of our future work, we plan to explore and identify a suitable measure capable of capturing style similarities independently of the image's content.

All three measures employed in this paper, namely subject-alignment, non-subject-disentanglement, and text-alignment, were calculated with the CLIP ViT-B/32 model. When calculating subject-alignment, we apply center alignment and image resizing to the subject segments before feeding them into the CLIP image encoder.

\setcounter{figure}{0}
\vspace{-0.2cm}
\section{Human evaluation}
\label{sec: Supple_human_evaluation}
\vspace{-0.2cm}
\begin{table}[h]
\resizebox{\columnwidth}{!}{%
\begin{tabular}{lcccc}
\toprule
Method                          & Subject-alignment & Non-subject-disentanglement & Text-alignment  & Overall         \\
\midrule
DreamBooth                      & 36.2\%   & 10.9\%                      & 7.8\%           & 15.5\%         \\
DreamBooth + SID                & \textbf{40.6\%}            & \textbf{68.8\%}             & \textbf{59.7\%} & \textbf{70.9\%} \\
Undecided         & 23.2\%     & 20.3\%                      & 32.5\%          & 13.6\%          \\
\bottomrule
\end{tabular}%
}
\caption{Human evaluation results.}
\label{tab: human evaluation}
\end{table}
\vspace{-0.4cm}

We conducted human evaluation involving 130 participants mainly recruited from our university community anonymously. 
The survey comprised 4 questions per subject, assessing subject-alignment, non-subject disentanglement, text-alignment, and the overall judgement considering all three aspects. 
Each participant evaluated 10 subjects, resulting in 40 problems, choosing between two images generated by DreamBooth and SID-integrated DreamBooth, respectively, and the order of the images was randomized.
In cases of difficulty, participants could opt for the ``Cannot Determine / Both Equally'' option. 
In \cref{tab: human evaluation}, the survey results show that our method was preferred over DreamBooth in all aspects, with particularly notable differences observed in \textit{text-alignment}, \textit{non-subject disentanglement}, and the overall judgement.
The results were also consistent with the results of our metric analysis in \cref{fig:multiple_reference_curve}, supporting our metric analysis results significantly.
% Notably, \cref{tab: human evaluation} shows that the survey results for subject-alignment, non-subject disentanglement, and text-alignment are consistent with the metric analysis in \cref{fig:multiple_reference_curve}, thereby supporting our measure-based analysis.
% \cref{tab: human evaluation} presents the survey results, revealing that our method significantly outperforms DreamBooth in all aspects.

\setcounter{figure}{0}
\section{Implementations and datasets in detail}

\subsection{Descriptions in style re-contextualization}
\label{sec: Supple_for style re-contextualization}
%In the case of style re-contextualization, we followed the approach of textual inversion~\cite{gal2022image} and custom diffusion~\cite{kumari2023multi}, using the format of ``A painting/cartoon in the style of [v] art".
%In accordance with this approach, we made slight modifications to the original instructions for generating SID as follows:
In style re-contextualization, we employed the phrase ``A painting/cartoon in the style of [v] art’’ as our baseline train description, which is also used in Textual Inversion~\cite{gal2022image} and Custom Diffusion~\cite{kumari2023multi} to capture the styles of reference images. For generating SID that extends the baseline description, we made a slight modification to the VLM instruction as follows:
\begin{itemize}
  \item[] \texttt{Describe the image in one sentence in detail. Please start the sentence with "A painting/cartoon in the style of art.". You should not describe the style of the painting/cartoon itself.}
\end{itemize}

\subsection{Model implementations}
\label{sec: Supple_implementation details}
\vspace{-0.2cm}
%In this section, we provide the implementation details of baseline models used in Sec. 4. For the optimization-based models, we employed the "stable-diffusion-2-1-base" as our backbone model, and for the encoder-based models, we utilized the backbone models employed during their pre-training phase without modification. 
In this section, we provide the implementation details of the baseline models employed in this paper. We selected ``stable-diffusion-2-1-base'' as the backbone models for DreamBooth~\cite{ruiz2023dreambooth}, SVDiff~\cite{han2023svdiff}, Custom Diffusion~\cite{kumari2023multi}, and Textual Inversion~\cite{gal2022image}. As for ELITE~\cite{wei2023elite} and BLIP-Diffusion~\cite{li2023blip}, we employed the backbone models used in their respective official GitHub repositories~\cite{ELITE_pytorch, BLIP_Diffusion_pytorch}.
All the generated images are sampled using the DDIM~\cite{song2020denoising} sampler, employing 50 steps and a guidance scale~\cite{ho2022classifier} of 7.5.

\textbf{DreamBooth~\cite{ruiz2023dreambooth}.} We implemented DreamBooth using the Hugging Face Diffusers library~\cite{von_Platen_Diffusers_State-of-the-art_diffusion}. The model was trained with a batch size of 1 and a learning rate of $1 \times 10^{-6}$ for 1000 iterations. We also trained a CLIP~\cite{radford2021learning} text encoder in conjunction with the U-Net~\cite{ronneberger2015u}.

\textbf{SVDiff~\cite{han2023svdiff}.} We implemented SVDiff using a third-party implementation~\cite{SVDiff_pytorch} due to the unavailability of official code. The model was trained for 1000 iterations with default hyperparameter settings.

\textbf{Custom Diffusion~\cite{kumari2023multi} \& Textual Inversion~\cite{gal2022image}.} We implemented Custom Diffusion and Textual Inversion using the Hugging Face Diffusers library~\cite{von_Platen_Diffusers_State-of-the-art_diffusion}, with default hyperparameter settings.
 
\textbf{ELITE~\cite{wei2023elite} \& BLIP-Diffusion~\cite{li2023blip}.} We implemented ELITE and BLIP-Diffusion using the official GitHub repositories~\cite{ELITE_pytorch, BLIP_Diffusion_pytorch}.

% \subsection{Evaluation metrics implementations}
% \textcolor{blue}{
% Since both subject-alignment and non-subject-disentanglement are calculated based on the subject segment, we utilized GroundedSAM~\cite{Grounded-SAM_Contributors_Grounded-Segment-Anything_2023} conditioned on the subject's class name to obtain the subject segment.
% The non-subject segment is derived by removing the region corresponding to the subject segment from the original image, and this is employed to achieve non-subject disentanglement.
% In the case of subject-alignment, since there are instances where the subject segment is situated in the corners or too small in the image, resizing and center alignment is performed. 
% An example of this can be observed in \cref{fig:C5 (b)}.(b).
% }

\subsection{Dataset details}

\begin{figure*}[t]
    \centering
    \includegraphics[width=0.9\textwidth]{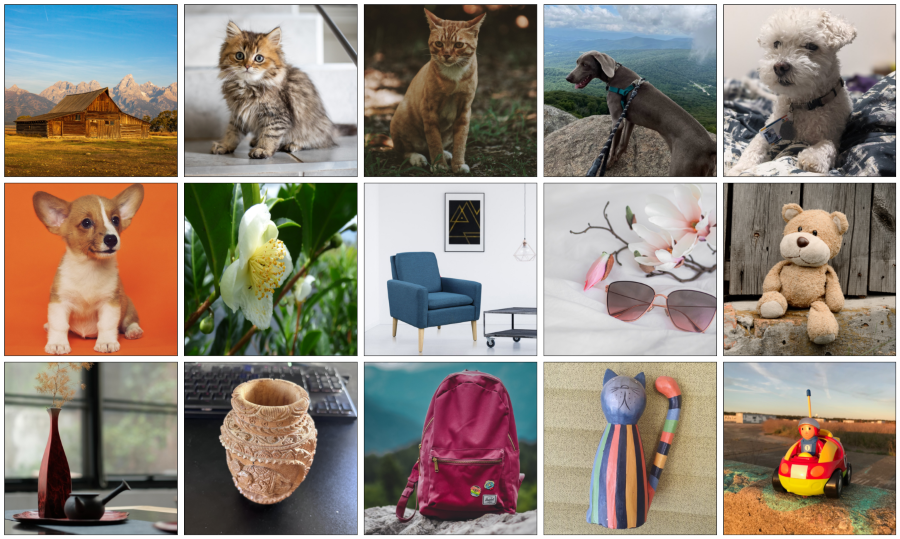}
    \caption{\textbf{Dataset used for quantitative analyses.} An example image of each subject is shown. The 15 subjects were chosen from three recent works: 7 from DreamBooth~\cite{ruiz2023dreambooth}, 7 from Custom Diffusion~\cite{kumari2023multi}, and 1 from Textual Inversion~\cite{gal2022image}.}
    \label{fig:images for quantitative}
\vspace{-0.35cm}
\end{figure*}

The quantitative analyses in our work are based on the dataset shown in \cref{fig:images for quantitative}. The 15 subjects in the dataset were chosen as the prominently utilized subjects in three recent text-to-image personalization  works~\cite{ruiz2023dreambooth, kumari2023multi, gal2022image}.
%We have chosen the 15 subjects prominently used in three previous works~\cite{ruiz2023dreambooth, kumari2023multi, gal2022image}, 
Specifically, 7 subjects were chosen from DreamBooth~\cite{ruiz2023dreambooth}: \texttt{[cat, dog3, dog6, pink sunglasses, vase, backpack, rc car]}, 7 subjects from Custom Diffusion~\cite{kumari2023multi}: \texttt{[barn, cat, dog, flower, chair, teddybear, wooden pot]}, and 1 from Textual Inversion~\cite{gal2022image}: \texttt{[cat toy]}.
The subjects cover a variety of categories including living animals (5 subjects), plants (1 subject), buildings (1 subject), toys (3 subjects), artistic containers (2 subjects), and three other objects (3 subjects). 
For each of these subjects, 25 generation prompts are used, and 20 images are generated per prompt with different random noise initializations, resulting in a total of 7500 generated images.

\subsection{VLM details}
In Sec. 3.2., we selected the multimodal GPT-4 as our choice for generating SID after evaluating three instruction-following VLMs, each with the following versions:

\begin{itemize}
    \item \textbf{BLIP-2~\cite{li2023blip2}}: ``blip2-opt-2.7b''~\cite{blip2-opt-2.7b}
    \item \textbf{LLaVA~\cite{liu2023visual}}: ``llava-v1-0719-336px-lora-merge-vicuna-13b-v1.3''~\cite{llava-v1-0719-336px-lora-merge-vicuna-13b-v1.3}
    \item \textbf{GPT-4~\cite{openai2023gpt4}}: Multi-modal GPT-4 provided in ChatGPT~\cite{ChatGPT}
\end{itemize}
% We have noticed that the GPT-4 Vision API was announced to be released after November 6th, 2023. However, as of now, we have not yet started using it, but we look forward to implementing it to simplify the SID generation process.
We have noticed that the GPT-4 Vision API was announced to be released after November 6th, 2023, and we look forward to using it to simplify the SID generation process.

Given the rapid evolution of VLMs, including the recent releases of GPT-4 Turbo with vision and LLaVA v1.5, we acknowledge that our analysis in Sec. 3.2. is limited to the versions mentioned above. Comparison results may be affected with the introduction of these new VLM versions or customized VLM instructions. We emphasize that our primary research focus is to identify a description format that can reduce undesired entanglement rather than the ease of generating train descriptions. We acknowledge that the optimal VLM for generating SID can change with the emergence of newer VLM versions at any moment.

\label{sec: Supple_datasets}

% Subject names: 
% DreamBooth(backpack, cat, dog3, dog6, pink_sunglasses, rc_car, vase)
% Custom Diffusion()

% barn (from Custom Diffusion)
% cat (from Custom Diffusion)
% cat (from DreamBooth)
% dog (from Custom Diffusion)
% dog3 (from DreamBooth)
% dog6 (from DreamBooth)
% flower (from Custom Diffusion)
% chair (from Custom Diffusion)
% pink_sunglasses (from DreamBooth)
% teddybear (from Custom Diffusion)
% vase (from DreamBooth)
% wooden_pot (from Custom Diffusion)
% backpack (from DreamBooth)
% cat toy (from textual Inversion)
% rc_car (from DreamBooth)
% 끝

% 혹시 Figure B.1. 순서대로 알려줄 수 있어? 첫 번째 행부터??
% subject 이름이 겹치는 군 ㅎㅎ 음...생각 중... OK

\setcounter{figure}{0}
\section{Limitations of SID}
\label{sec: Supple_limitations of SID}
\begin{figure*}[t]
\centering
\begin{subfigure}[c]{0.8\textwidth}
    \centering
    \includegraphics[width=\columnwidth]{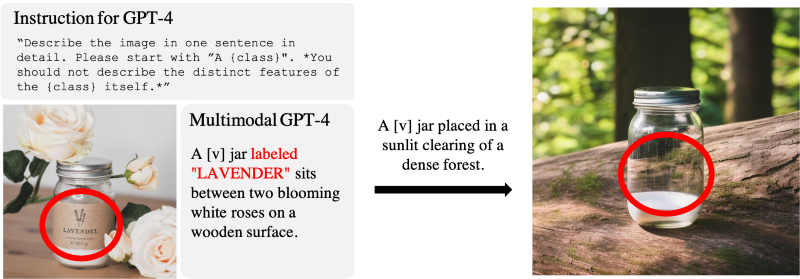}
    \caption{Jar}
\end{subfigure}

\begin{subfigure}[c]{0.8\textwidth}
    \centering
    \includegraphics[width=\columnwidth]{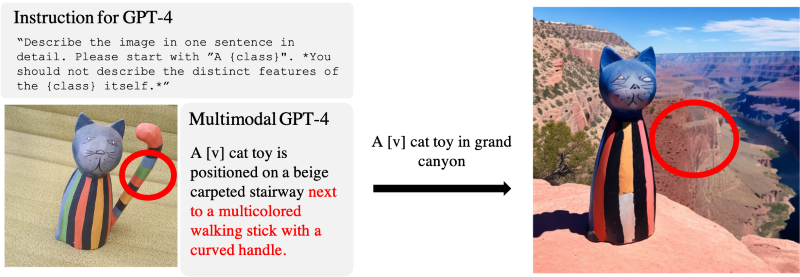}
    \caption{Cat toy}
\end{subfigure}
\vspace{-0.3cm}
\caption{
% VLM failure cases
\textbf{Imperfection of GPT-4.} In (a), the description generated by GPT-4 includes an informative specification of the label attached to the jar, leading to the omission of the subject's label in the generated image. 
In (b), the description depicts the tail of the cat toy as a walking stick, resulting in the omission of the tail in the generated image.
}
\label{fig:failure cases of GPT-4}
% \vspace{-0.3cm}
\end{figure*}

\begin{figure*}[t]
    \centering
    \includegraphics[width=\textwidth]{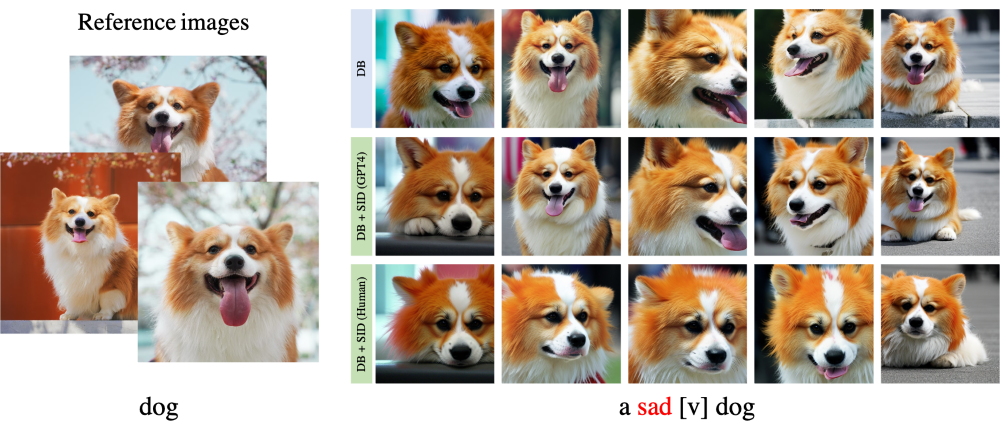}
    \vspace{-0.3cm}
    \caption{
    \textbf{Selectively describing facial expression.} When the SID generated by GPT-4 lacks information about facial expression, the SID-integrated model fails to modify the subject's facial expression, as depicted in the middle row. By manually designing SID to describe the subject’s facial expression, the SID-integrated model successfully disentangles the subject identity from the facial expression, as demonstrated in the last row. The manually designed SID is ``a [v] dog with a joyful and open-mouthed expression, with its tongue hanging out."
    % DreamBooth tends to suffer from altering the subject's expression according to the generation prompt. Additionally, integration with SIDs generated by GPT-4 did not yield significant improvements. We attributed this to the lack of information regarding the subject's facial expression in the generated SIDs. Therefore, in the last row, we conducted experiments by manually describing the subject's expression and observed successful outcomes.
    % Manually designed text description of facial expression is as follows: ``a [v] dog with a joyful and open-mouthed expression, with its tongue hanging out."
    }
    \vspace{-0.3cm}
    \label{fig:rebuttal_facial}
\end{figure*}

\subsection{Failure cases with GPT-4}
\label{sec: Limitation_GPT_4_failure}
While GPT-4~\cite{openai2023gpt4} generally generates descriptions that closely match the provided instructions, occasional failures do occur. In \cref{fig:failure cases of GPT-4}, we report two instances where GPT-4 does not generate prompts that align closely with the properties of SID.

\subsection{Selectively describing facial expression}
\label{sec: Limitation_facial_expression}
We observed that SID-integrated models often face difficulties in altering the subject's facial expression according to the generation prompt, especially when the reference images display strong facial expressions.
This challenge arises due to the lack of text descriptions of the subject's facial expressions in the VLM-generated SID, resulting in undesired entanglement of the facial expressions in the subject embedding.
We discovered that this challenge can be addressed by incorporating text descriptions of the facial expressions into the SID.
Specifically, this can be achieved by modifying the instructions of VLM for SID or manually adding descriptions of the facial expressions into SID.
% , the challenge in feature disentanglement can be handled.
For example, in \cref{fig:rebuttal_facial}, we demonstrated that manually adding descriptions of the subject's facial expression can resolve the undesired feature entanglement issue, successfully altering the subject's facial expression.
% \cref{fig:rebuttal_facial}.
% While it would be best to automatically generate SIDs describing every undesirable feature, devising a general VLM instruction to achieve this has posed challenges.
% Therefore, if we aim to disentangle features that are missing from the SID, we can design the desired SID by either modifying the instructions of VLM or manually adding descriptions for the respective features.
% SID is designed to refrain from describing the features of the subject and instead focus on detailing other undesired objects.
% Such selective design aims to disentangle other undesirable information from subject embedding while preserving the subject's identity.
% However, there exist some undesirable features which are inherently ingrained within the subject itself but do not affecting subject's identity, such as pose and facial expression.
% As these undesirable features are inherent to the subject itself, VLMs, which are instructed to refrain from describing the subject, often encounter difficulty in generating relevant descriptions.

\setcounter{figure}{0}
\section{Additional experiment results}
In this section, we provide additional experiment results to augment those presented in the main text. Additionally, we conduct qualitative comparisons between DreamBooth~\cite{ruiz2023dreambooth} and its SID-integrated counterpart in scenarios with varying levels of biases.

\subsection{Four description cases}
\label{sec: Supple_four description cases}
\begin{figure*}[t]
    \centering
    \includegraphics[width=0.75\textwidth]{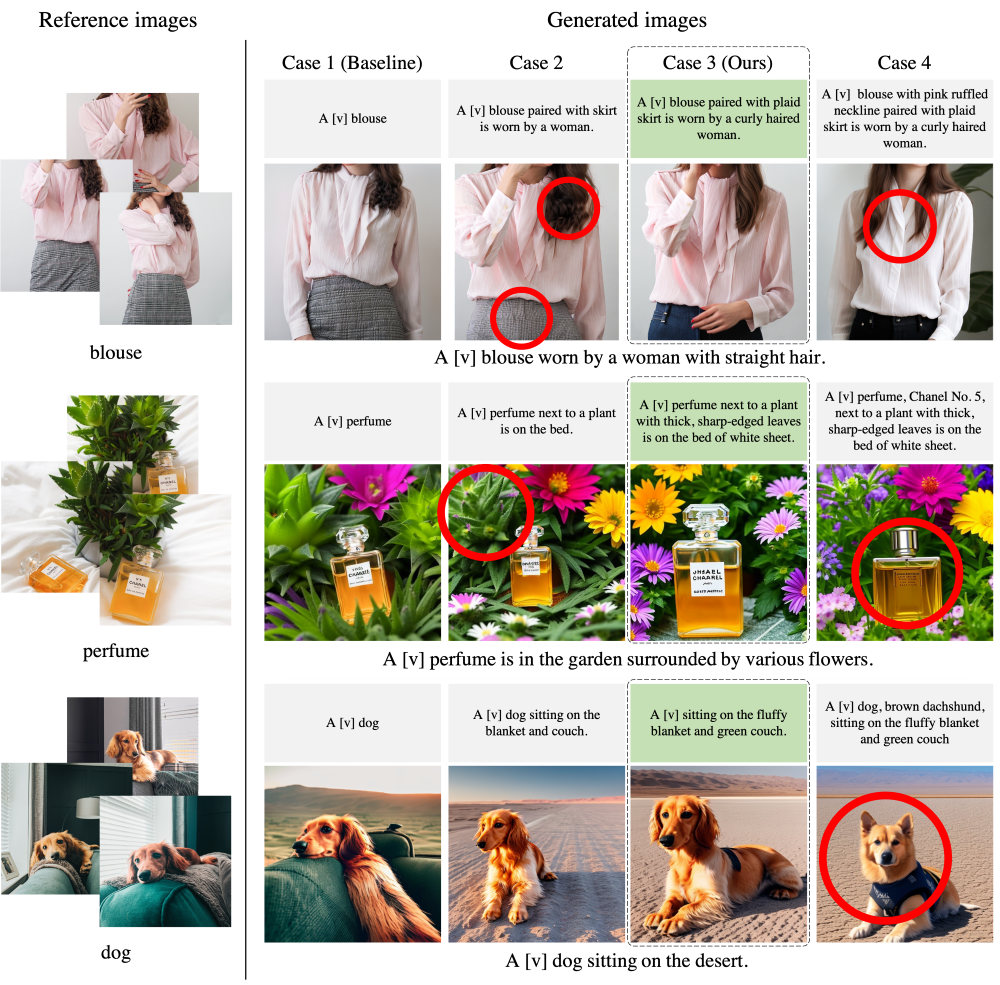}
    \vspace{-0.2cm}
    \caption{\textbf{Additional examples for comparing the four cases of descriptions.} Case 2 shows a decent generation with occasional entanglements such as (top row) curly hair, not aligned with generation prompt, and gray plaid pattern and (middle row) spiky plant. Case 4 demonstrates high text-alignment but significantly falls short in subject preservation. Case 3 (Ours) successfully achieves the desired qualities: text-alignment, subject-alignment, and non-subject-disentanglement.}
    \label{fig:additional_description_cases}
    \vspace{-0.1cm}
\end{figure*}

To further validate the effectiveness of Case 3 (Ours) among the four description cases defined in \cref{tab:prompt_ablation}, we conducted a quantitative comparison. The results are reported in \cref{tab:description_cases} and additional qualitative comparisons are shown in \cref{fig:additional_description_cases}. In \cref{tab:description_cases}, Case 1 (Baseline) exhibits high subject alignment but lacks in text-alignment and non-subject disentanglement, indicating undesired embedding entanglement. On the other hand, Case 4 demonstrates optimal text-alignment and high non-subject disentanglement but significantly falls short in subject-alignment, as also evident in \cref{fig:additional_description_cases}. In contrast, both Case 2 and 3 achieve decent scores across all three measures, with Case 3 (Ours) outperforming Case 2 in all aspects. To conclude, Case 3 (Ours) achieves the best subject-alignment and non-subject-disentanglement, along with near-best text-alignment scores, which agree with the additional qualitative comparisons in \cref{fig:additional_description_cases}. 

% [2024-03-22]
% \textcolor{orange}{
% For a more comprehensive understanding of each case, in \cref{fig:rebuttal_attention}, we additionally visualized the cross-attention map for each case.
% In Cases 1 and 2, the attention maps of [v] undesirably focus on the nearby-object, the `purse.' Conversely, Cases 3 and 4 exclusively concentrate on the subject—the `perfume.' However, Case 4 yields an incorrect perfume.
% This analysis reinforces our conclusion.}

%We show quantitative comparison of four cases of descriptions in \cref{tab:description_cases}, and additional qualitative results in
%We show additional qualitative comparison of four cases of descriptions in \cref{fig:additional_description_cases}.
% \textcolor{blue}{Figure 3}

\vspace{-0.025cm}
\subsection{Instruction-following VLMs}
\label{sec: Supple_instruction-following VLMs}
\vspace{-0.03cm}
To support the explanations presented in Sec. 3.2., we present additional comparisons of the three instruction-following VLMs in \cref{fig:additional_VLM_comparison}. Additionally, we provide quantitative comparisons among the three VLMs to assess whether the multi-modal GPT-4~\cite{openai2023gpt4} outperforms the other two VLMs, namely LLaVA~\cite{liu2023visual} and BLIP-2~\cite{li2023blip2}, in terms of text-alignment, subject-alignment, and non-subject-disentanglement. The results are shown in \cref{tab:VLM_comparison}, demonstrating that GPT-4 excel over the other two VLMs in all three measures.

\vspace{-0.025cm}
% [2024-03-22 교수님 미팅 후 추가]
\subsection{Cross-attention map analysis}
\begin{figure*}[h]
    \centering
    \includegraphics[width=0.75\textwidth]{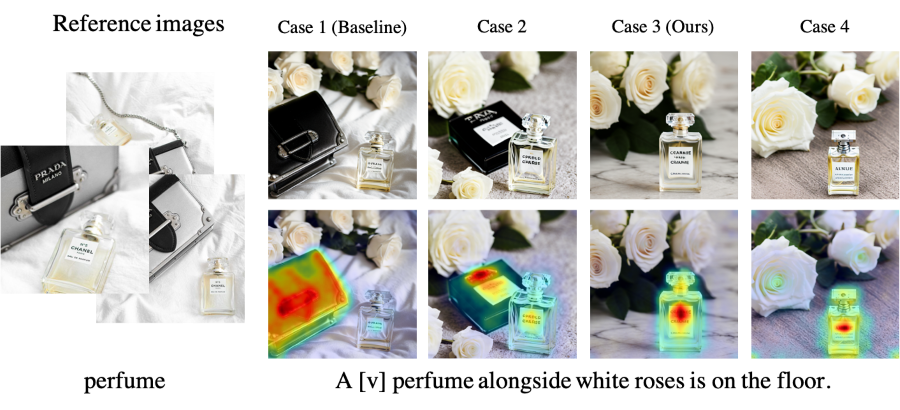}
    \vspace{-0.4cm}
    \caption{\textbf{Cross-attention maps of the identifier [v] for the four cases of train descriptions from \cref{fig:prompt_ablation}, top row.}
    In Cases 1 and 2, attention is spread towards non-subject parts.
    However, in Cases 3 and 4, thanks to the informative specification of the non-subject, this spreading of attention is highly reduced.
    } 
    \label{fig:rebuttal_attention}
\end{figure*}

\begin{figure*}[t]
    \centering
    \includegraphics[width=\textwidth]{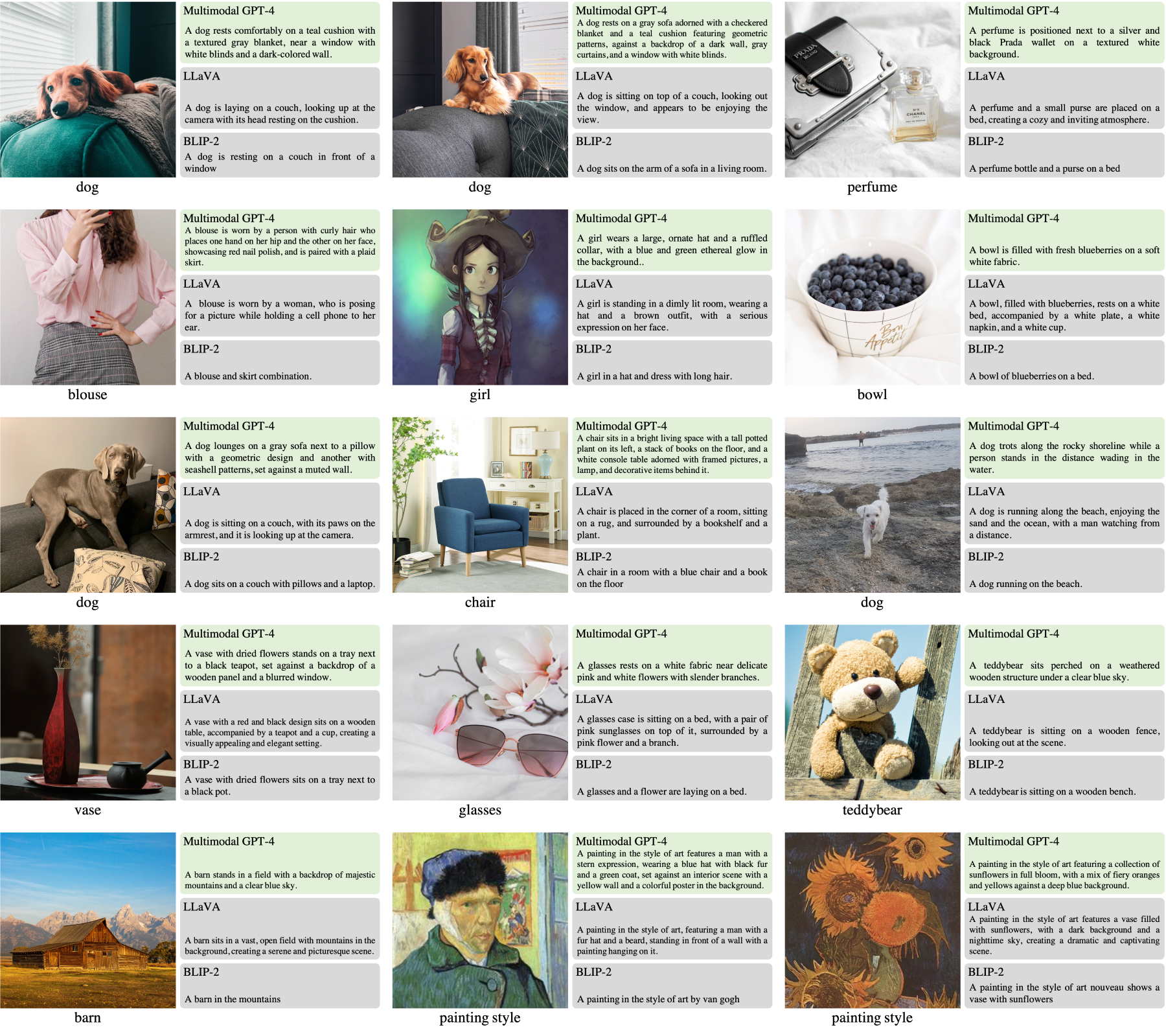}
    \caption{
    \textbf{Additional examples for comparing the three instruction-following VLMs for generating SID.}
    LLaVA tends to include informative specifications of the subject itself or fall short in providing sufficient specifications of the undesired objects.
    BLIP-2 exhibits limitations in achieving a thorough identification of the undesired objects. Even when it successfully identifies an undesired object, it tends to generate simple captions without informative specifications.
    Compared to the other two, GPT-4 excels in generating captions that satisfy our instructions.
    % BLIP-2 tends to lack sufficient identification of a diverse range of undesired objects, and even when it does, it tends to generate simple captions without informative specifications.
    %LLaVA tends to include informative specifications of the subject itself or fall short in providing sufficient specifications of undesired objects.
    }
    % \label{fig:vs_Neg_Seg}
    \label{fig:additional_VLM_comparison}
\end{figure*}

\vspace{-0.03cm}
To extend the analysis presented in \cref{sec:Analysis_cross}, we conducted the cross-attention map analysis on the four cases of train descriptions detailed in \cref{fig:prompt_ablation}, top row. In \cref{fig:rebuttal_attention}, our analysis focused on the nearby-object bias, which can similarly be applied to other biased scenarios. In Cases 1 and 2, the identifier [v] erroneously focuses on the nearby-object, ‘the purse’, in each generated image. Conversely, Cases 3 and 4 demonstrate a notable improvement in focusing on the subject, ‘the perfume’. The primary distinction between these two groups of cases lies in the inclusion of informative specifications related to the non-subjects. This indicates that incorporating informative specifications can reduce the undesired focus or generation of the non-subjects. However, it is crucial to include informative specification \emph{selectively}, as demonstrated in Case 4, where the inclusion of the informative specification related to the subject destroys its identity.

%We show quantitative comparison of three instruction-following VLMs in \cref{tab:VLM_comparison}, and additional qualitative results in \cref{fig:additional_VLM_comparison}.

% \clearpage
% \vspace{-0.2cm}
\vspace{-0.025cm}
\subsection{Enhancement by SID}
\label{sec: Supple_enhancement by SID}
\vspace{-0.03cm}
We present comprehensive generation results of Fig. 5 in \cref{fig:C3 (a)} (a), (b), (c), and (d), where the generation outcomes are presented without any selection. Additional experimental results can be found in \cref{fig:C3 (e)} (e), (f), (g), and (h). We observed that, in the context of undesired embedding disentanglement, SVDiff stands out among other models. In case of Textual Inversion, the baseline model exhibits a low level of subject alignment, resulting in the
SID-integrated model encountering a similar challenge.

% \clearpage
\subsection{Enhancement for a single reference image}
\label{sec: Supple_enhancement for a single reference image}
\vspace{-0.05cm}
We present comprehensive and additional results of Fig. 6 in \cref{fig:C4 (a)} (a), (b), (c), $\dots$, (h).

% \clearpage
\subsection{Negative prompt and segmentation}
\label{sec: Supple_negative prompt and segmentation}
\vspace{-0.05cm}
We present comprehensive and additional results of Fig. 10 in \cref{fig:C5 (a)} (a) and (b).

% \clearpage
\subsection{Highly, moderately, and low-biased scenarios}
\label{sec: varying levels of biases}
\vspace{-0.05cm}
We compare DreamBooth~\cite{ruiz2023dreambooth} with its SID-integrated counterpart in scenarios with varying levels of biases: high, moderate, and low. The generation results for each scenario can be found in \cref{fig:C6 (a)} (a), (b), and (c).

\begin{table}[t]
\resizebox{\columnwidth}{!}{%
\begin{tabular}{lccc}
\toprule
Descriptions      & \begin{tabular}[c]{@{}c@{}}Text\\ alignment\end{tabular} & \begin{tabular}[c]{@{}c@{}}Subject \\ alignment\end{tabular} & \begin{tabular}[c]{@{}c@{}}Non-subject \\ disentanglement\end{tabular} \\
\midrule
Case 1 (Baseline) & 0.290                                                    & \textbf{0.686}                                               & 0.299                                                                  \\
Case 2            & 0.311                                                    & 0.685                                                        & 0.342                                                                  \\
Case 3 (Ours)     & 0.317                                                    & \textbf{0.686}                                               & \textbf{0.352}                                                         \\
Case 4            & \textbf{0.325}                                           & 0.659                                                        & 0.348                                     \\
\bottomrule
\end{tabular}%
}
\vspace{-0.2cm}
\caption{\textbf{Quantitative comparison of four description cases.} 
We have performed a quantitative analysis of the four cases of descriptions listed in Tab. 1.
%For each case, the choice of train description follows the guidelines in Tab. 1.
Case 3 (Ours) achieves the best performance for subject alignment and non-subject disentanglement. It also significantly improves text-alignment when compared to the Case 1 (Baseline).}
\label{tab:description_cases}
\vspace{-0.2cm}
\end{table}

\begin{table}[]
\centering
\resizebox{\columnwidth}{!}{%
\begin{tabular}{lccc}
\toprule
VLMs  & \begin{tabular}[c]{@{}c@{}}Text \\ alignment\end{tabular} & \begin{tabular}[c]{@{}c@{}}Subject \\ alignment\end{tabular} & \begin{tabular}[c]{@{}c@{}}Non-subject\\ disentanglement\end{tabular} \\
\midrule
GPT-4~\cite{openai2023gpt4}  & \textbf{0.317}                                            & \textbf{0.684}                                               & \textbf{0.354}                                                        \\
LLaVA~\cite{liu2023visual} & 0.311                                                     & 0.681                                                        & 0.340                                                                 \\
BLIP-2~\cite{li2023blip2} & 0.311                                                     & 0.683                                                        & 0.343                                                                \\
\bottomrule
\end{tabular}%
}
\vspace{-0.2cm}
\caption{\textbf{Quantitative comparison of instruction-following VLMs.} The multi-modal GPT-4 demonstrates superior performance in all three measures when used for generating SIDs.}
\label{tab:VLM_comparison}
\vspace{-0.55cm}
\end{table}
\begin{figure*}[t]
    \centering
    \begin{subfigure}{\textwidth}
    \includegraphics[width=\textwidth]{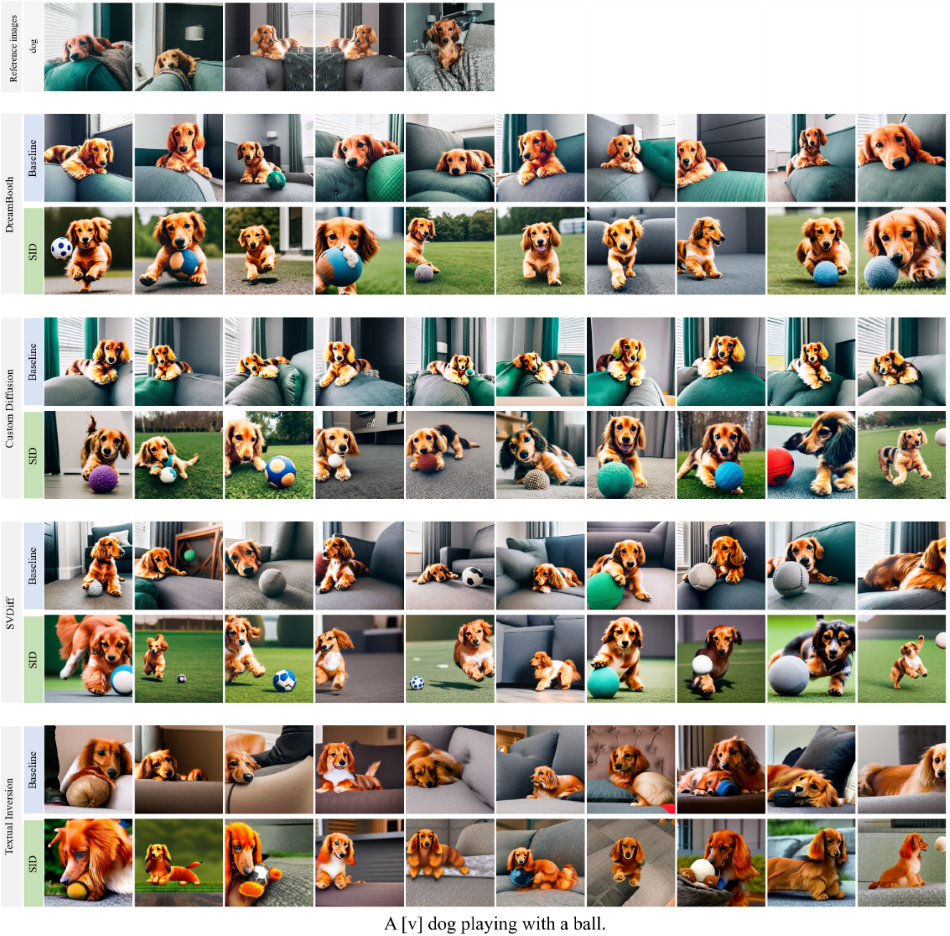}
    \end{subfigure}
    \caption{(a) \textbf{Enhancement by SID -- additional example \#1.} 
    SID-integration effectively resolves the entanglement problem of indoor background (background bias).
    % For textual inversion, the baseline model exhibits a low level of subject alignment, resulting in the SID-integrated model encountering a similar challenge.}
    % The images produced by the baseline model consistently display entanglement with indoor backgrounds, while the SID-integrated models generate backgrounds that better complement the given generation prompt, such as a grassy field. For textual inversion, the baseline model exhibits a low level of subject alignment, resulting in the SID-integrated model encountering a similar challenge.
    }
    \label{fig:C3 (a)}
\end{figure*}

\begin{figure*}[t]\ContinuedFloat
    \centering
    \begin{subfigure}{\textwidth}
    \includegraphics[width=\textwidth]{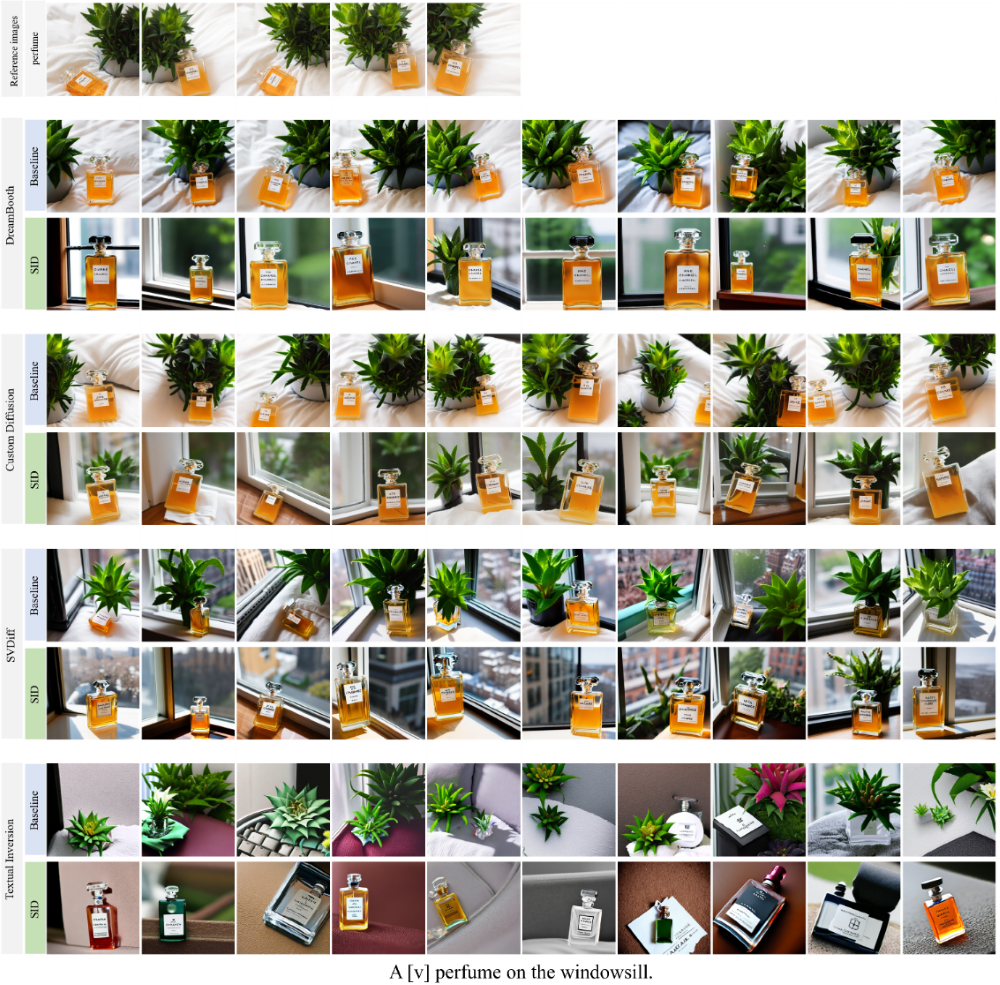}
    \end{subfigure}
    \caption{(b) \textbf{Enhancement by SID -- additional example \#2.} 
    SID-integration effectively resolves the entanglement problems of potted plant and white sheet background (nearby-object bias).
    % Images produced by baseline models frequently feature a potted plant and a white sheet elements commonly found in reference images that may not align with the provided generation prompt. 
    % Applying SID could diminish the presence of undesired objects and produce images that more closely match the given generation prompt.
    % We found that SVDiff stands out among other models, demonstrating a capacity to generate images that align well with the generation prompt, even in the baseline. This is evident in its effective creation of elements such as a windowsill, setting it apart in terms of disentangling undesired embedding.
    }
\end{figure*}

\begin{figure*}[t]\ContinuedFloat
    \centering
    \begin{subfigure}{\textwidth}
    \includegraphics[width=\textwidth]{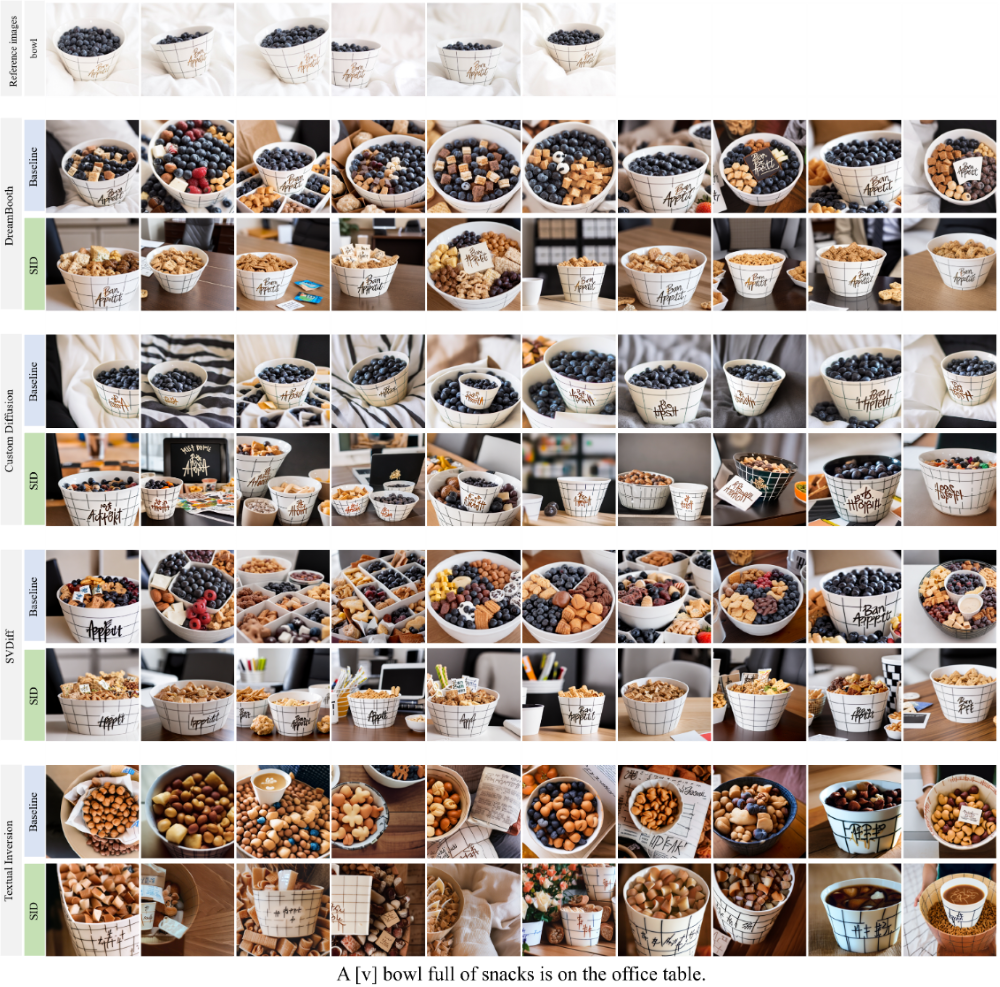}
    \end{subfigure}
    \caption{(c) \textbf{Enhancement by SID -- additional example \#3.} 
    SID-integration effectively resolves the entanglement problem of filled-in blueberries (tied-object bias).
    % The images generated by the baseline models exhibit entanglement with filled-in blueberries and produce an unnatural background.
    % In contrast, SID integrated models successfully removed blueberry, filled in snacks, and generated an appropriate background for the generation prompt.
    % In case of Textual Inversion, both the baseline model and the SID integrated model exhibit poor subject alignment.
    }
\end{figure*}

\begin{figure*}[t]\ContinuedFloat
    \centering
    \begin{subfigure}{\textwidth}
    \includegraphics[width=\textwidth]{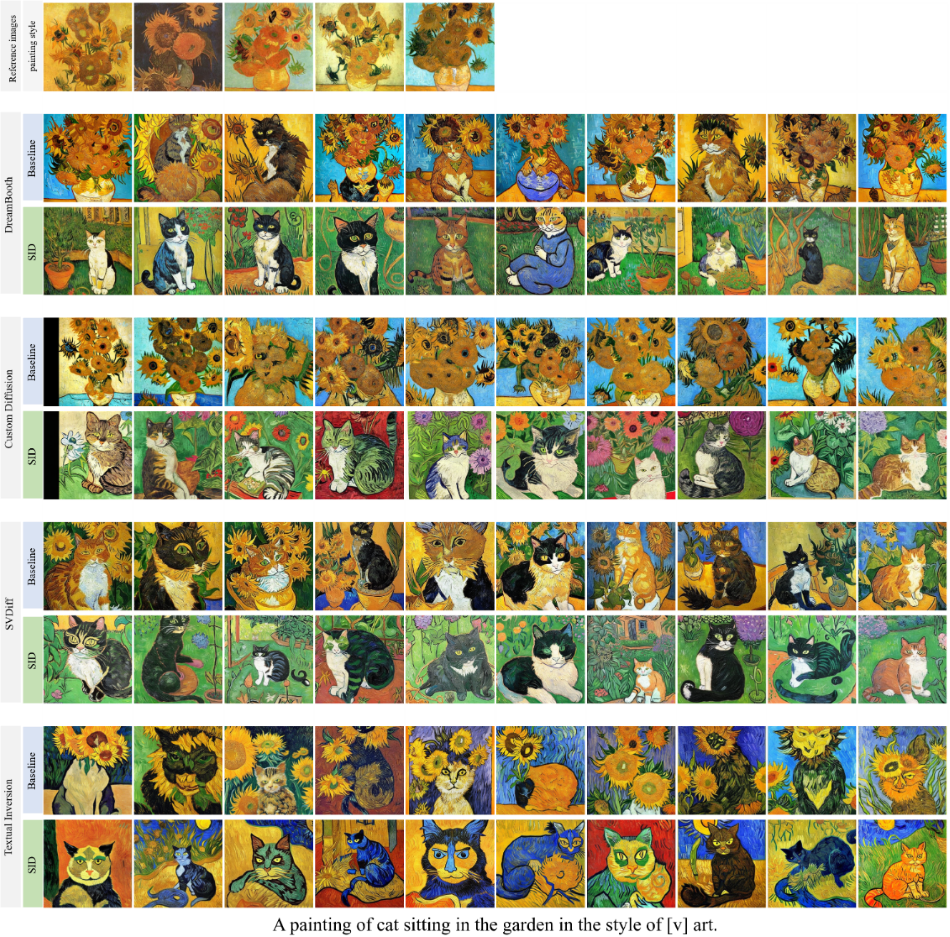}
    \end{subfigure}
    \caption{(d) \textbf{Enhancement by SID -- additional example \#4.} 
    SID-integration effectively resolves the entanglement problem of sunflowers (substance bias).
    }
    % The images produced by the baseline models display entanglement with the sunflower substance. 
    % Among these baselines, SVDiff demonstrates the lowest level of undesired embedding entanglement.
    % The integration of SID allows for a substantial reduction in such entanglement.}
\end{figure*}

\begin{figure*}[t]\ContinuedFloat
    \centering
    \begin{subfigure}{\textwidth}
    \includegraphics[width=\textwidth]{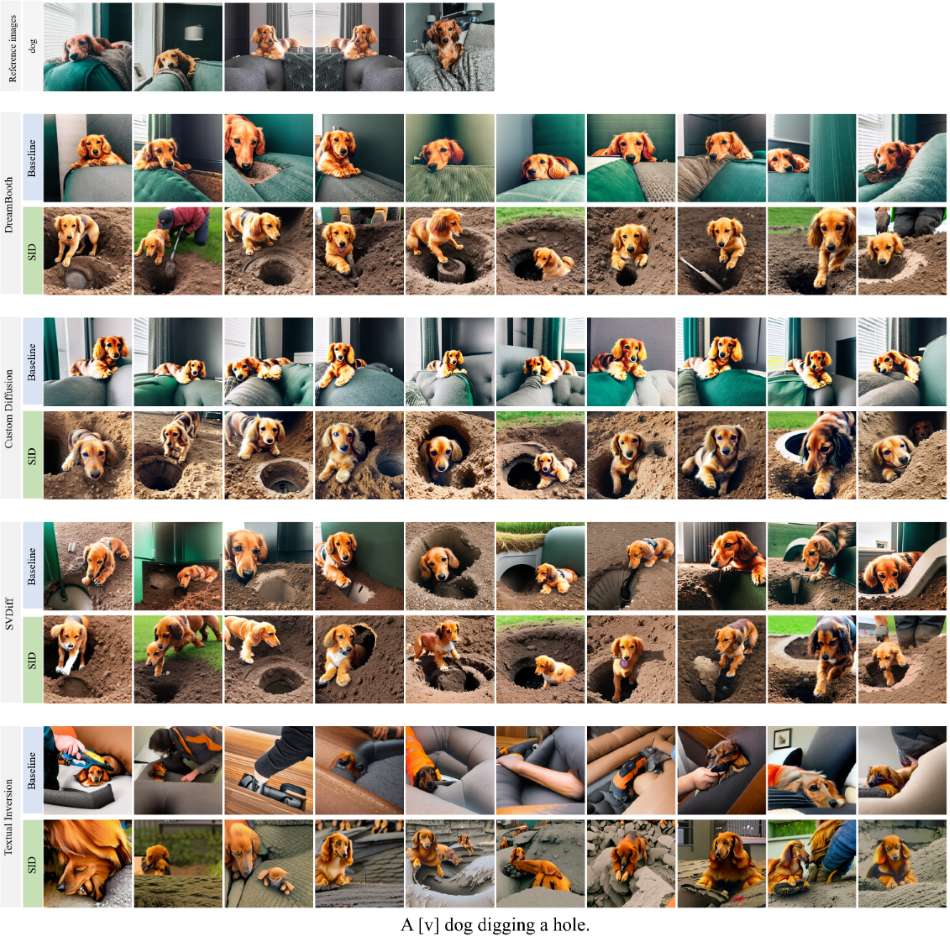}
    \end{subfigure}
    \caption{(e) \textbf{Enhancement by SID -- additional example \#5.} 
    SID-integration effectively resolves the entanglement problem of indoor background (background bias).
    }
    \label{fig:C3 (e)}
\end{figure*}

\begin{figure*}[t]\ContinuedFloat
    \centering
    \begin{subfigure}{\textwidth}
    \includegraphics[width=\textwidth]{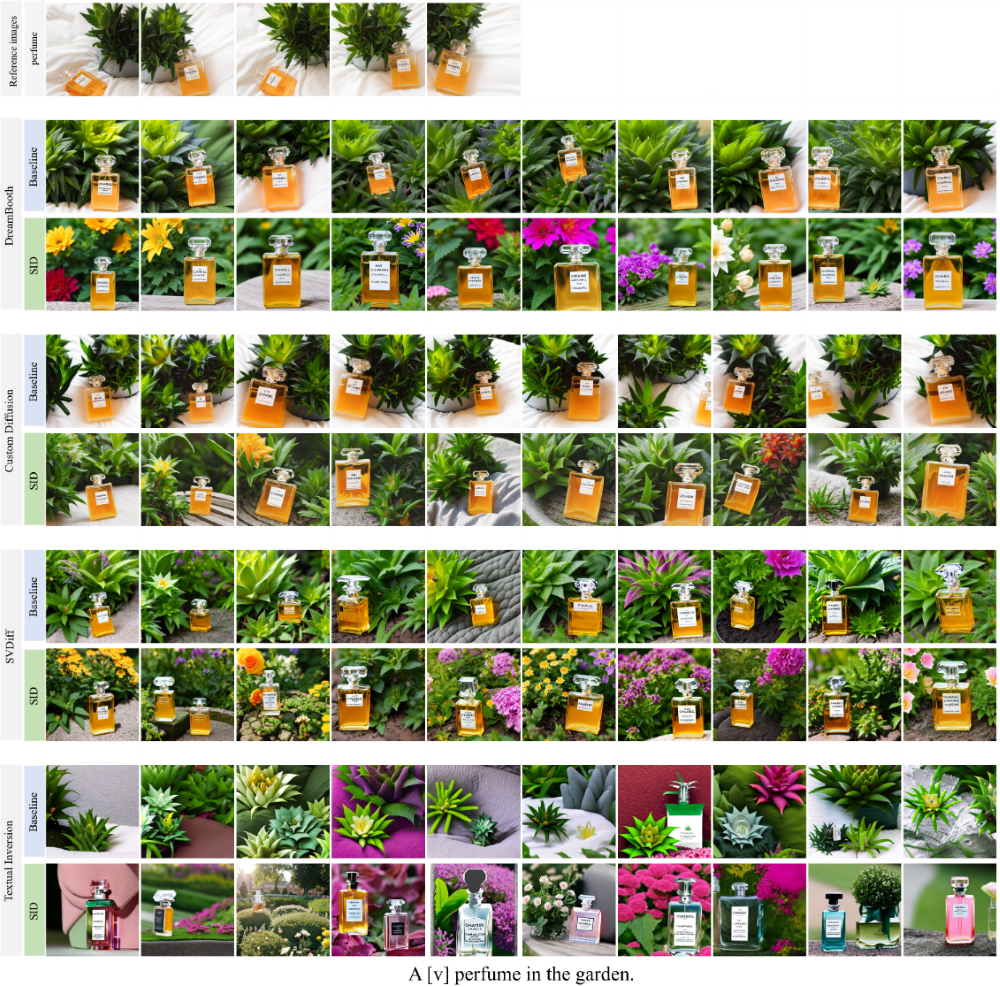}
    \end{subfigure}
    \caption{(f) \textbf{Enhancement by SID -- additional example \#6.} 
    SID-integration effectively resolves the entanglement problem of a plant with sharp-edged leaves (nearby-object bias).
    }
\end{figure*}

\begin{figure*}[t]\ContinuedFloat
    \centering
    \begin{subfigure}{\textwidth}
    \includegraphics[width=\textwidth]{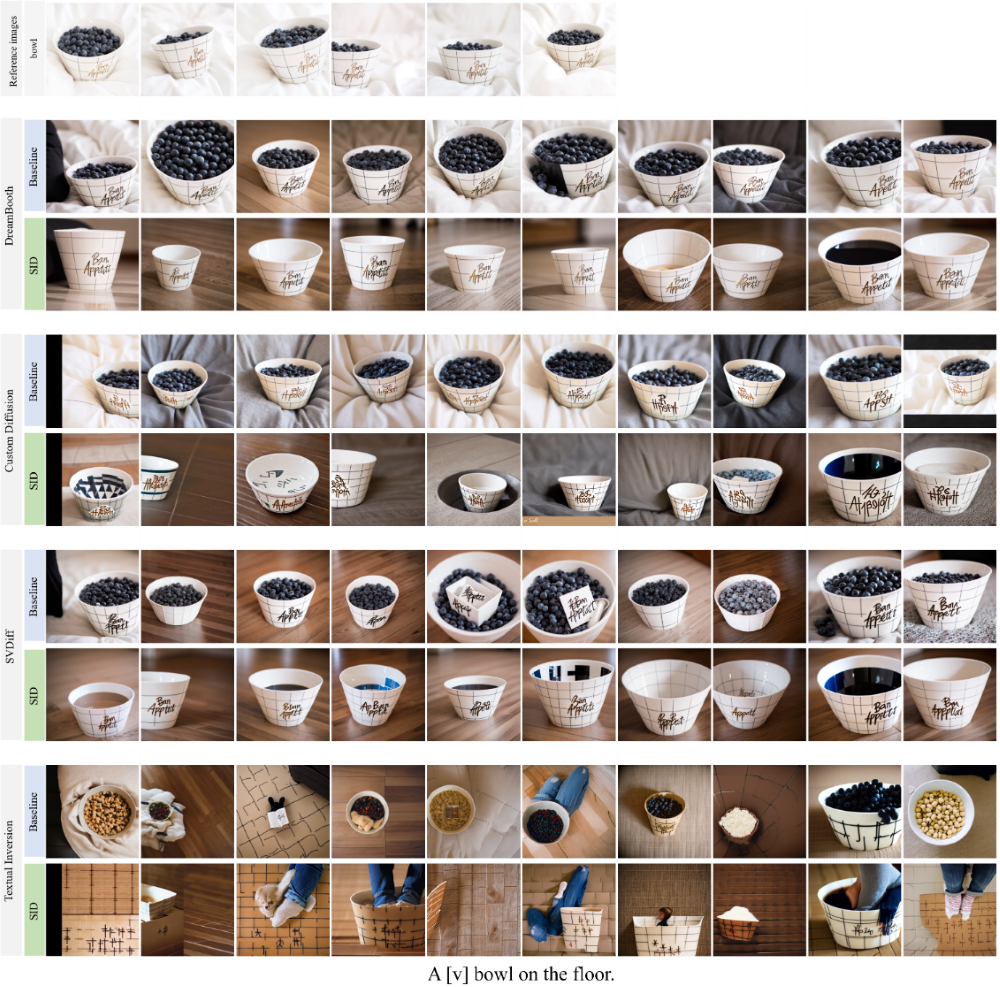}
    \end{subfigure}
    \caption{(g) \textbf{Enhancement by SID -- additional example \#7.} 
    SID-integration effectively resolves the entanglement problem of filled-in blueberries and a white sheet background (tied-object bias and background bias).
    }
\end{figure*}

% \begin{figure*}[t]\ContinuedFloat
%     \centering
%     \begin{subfigure}{\textwidth}
%     \includegraphics[width=\textwidth]{Figures/Appendix/Multi_vs_others/Multi_vs_others_8.png}
%     \end{subfigure}
%     \caption{(h) \textbf{Enhancement by SID -- additional example \#8.} 
%     SID-integration effectively resolves the entanglement problem of a brown bear (substance bias).
%     }
% \end{figure*}
% \begin{figure*}[t]
%     \centering
%     \includegraphics[width=\textwidth]{Figures/Appendix/One_vs_encoder/One_vs_Encoder_1.png}
%     \caption{aa}
% \end{figure*}
% \begin{figure*}[t]
%     \centering
%     \includegraphics[width=\textwidth]{Figures/Appendix/One_vs_encoder/One_vs_Encoder_2.png}
%     \caption{aa}
% \end{figure*}

\begin{figure*}[t]
    \centering
    \begin{subfigure}{\textwidth}
    \includegraphics[width=\textwidth]{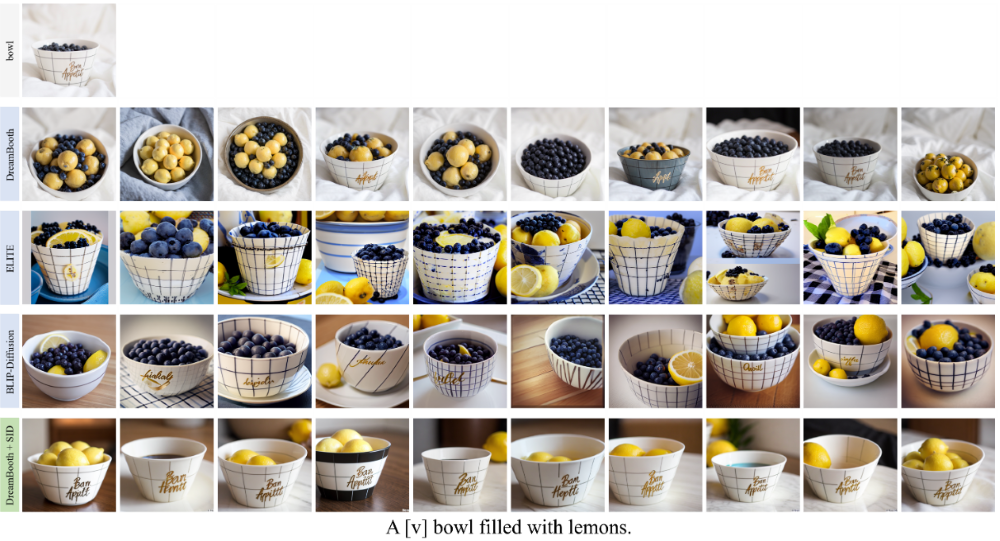}
    \end{subfigure}
    \caption{
    (a) \textbf{Enhancement by SID for a single reference image -- additional example \#1.} 
    DreamBooth and encoder-based models fail at removing blueberries from the bowl. Furthermore, they often fail at filling the bowl with lemons.
    }
    \label{fig:C4 (a)}
\end{figure*}

\begin{figure*}[t] \ContinuedFloat
    \centering
    \begin{subfigure}{\textwidth}
    \includegraphics[width=\textwidth]{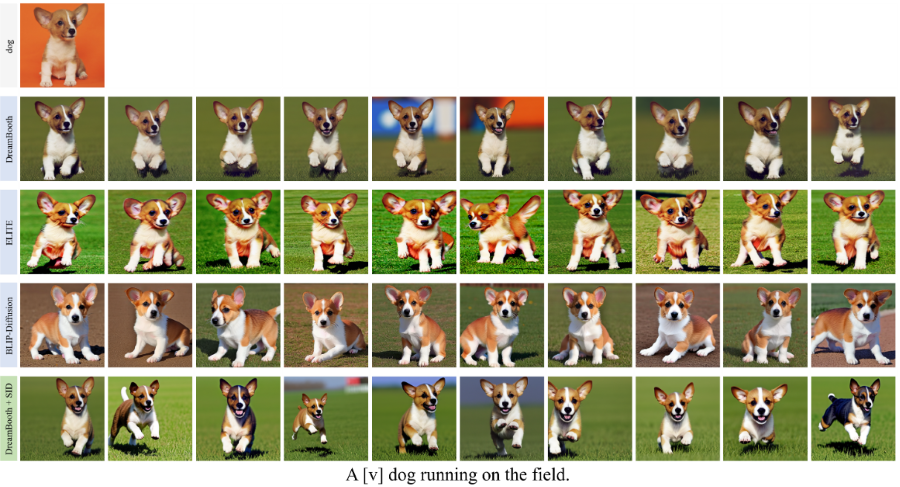}
    \end{subfigure}
    \caption{
    (b) \textbf{Enhancement by SID for a single reference image -- additional example \#2.} 
    DreamBooth and encoder based models struggle with generating images of a dog running in diverse poses (pose bias).
    }
\end{figure*}

\begin{figure*}[t] \ContinuedFloat
    \centering
    \begin{subfigure}{\textwidth}
    \includegraphics[width=\textwidth]{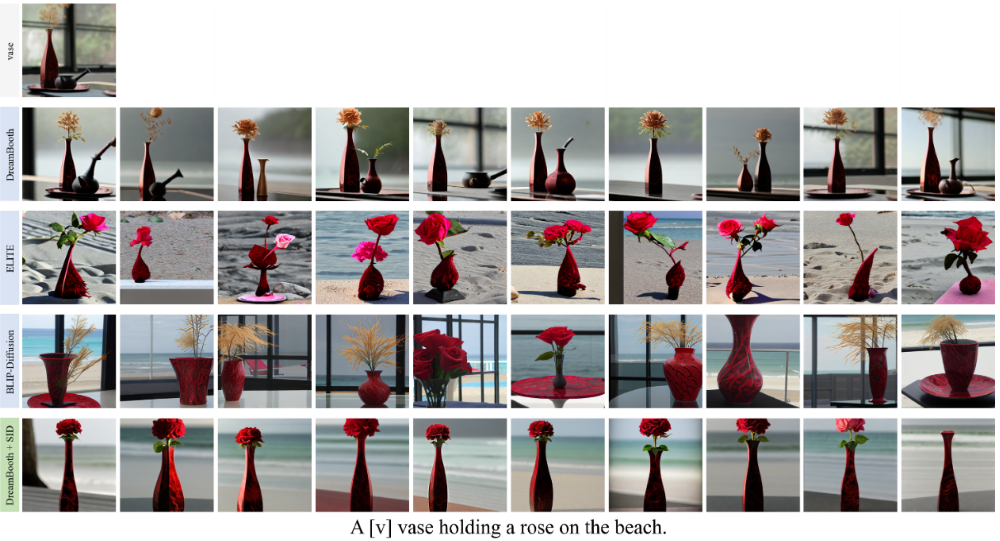}
    \end{subfigure}
    \caption{
    (c) \textbf{Enhancement by SID for a single reference image -- additional example \#3.} 
    DreamBooth not only generates a vase but also the nearby teapot and saucer. 
    Encoder-based models struggle to preserve the identity of the vase, possibly because of the infrequent occurrences of vases during the encoder pre-training.
    }
\end{figure*}

\begin{figure*}[t] \ContinuedFloat
    \centering
    \begin{subfigure}{\textwidth}
    \includegraphics[width=\textwidth]{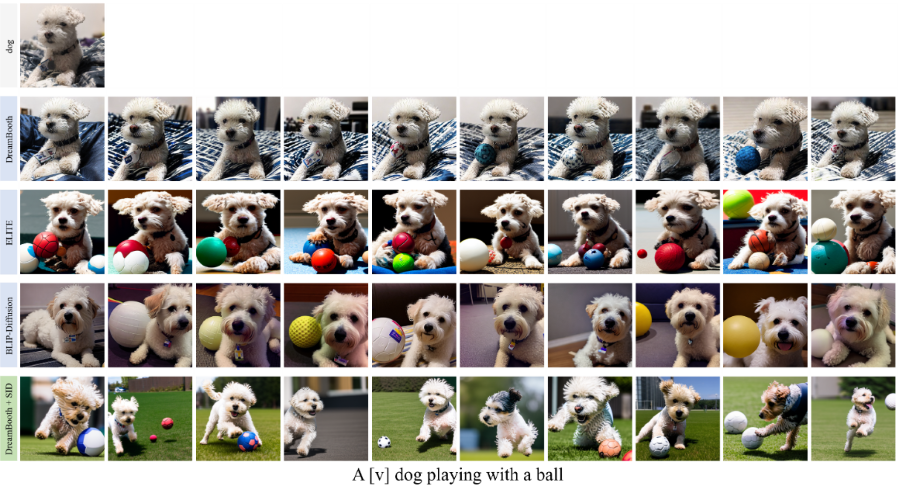}
    \end{subfigure}
    \caption{
    (d) \textbf{Enhancement by SID for a single reference image -- additional example \#4.} 
    DreamBooth and encoder based models struggle with generating images of a dog playing with a ball in diverse poses (pose bias).
    }
\end{figure*}

\begin{figure*}[t] \ContinuedFloat
    \centering
    \begin{subfigure}{\textwidth}
    \includegraphics[width=\textwidth]{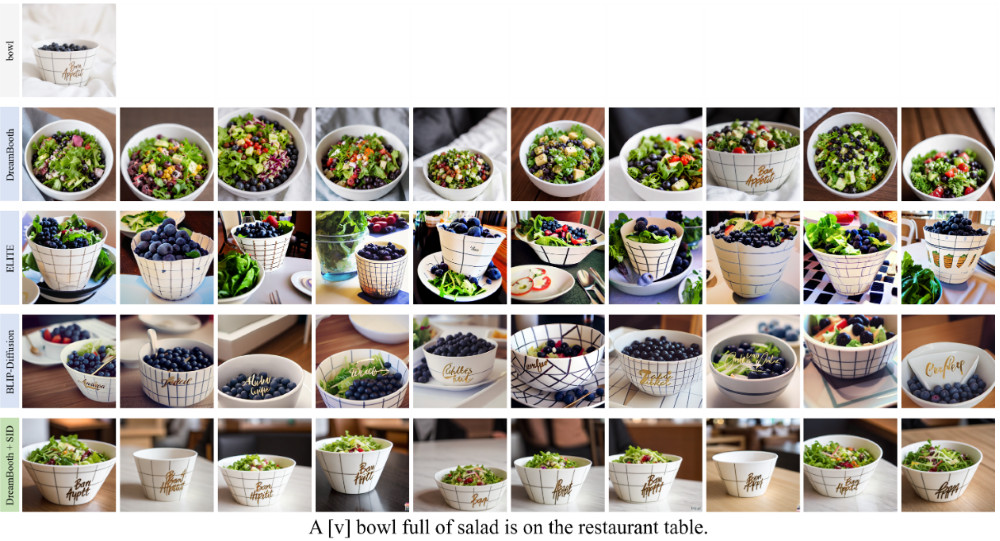}
    \end{subfigure}
    \caption{
    (e) \textbf{Enhancement by SID for a single reference image -- additional example \#5.} 
    DreamBooth struggles with preserving the identity of the bowl and also with generating a background that is aligned with the generation prompt. 
    Encoder-based models face challenges in removing blueberries from the bowl and replacing them with salad.
    }
\end{figure*}

\begin{figure*}[t] \ContinuedFloat
    \centering
    \begin{subfigure}{\textwidth}
    \includegraphics[width=\textwidth]{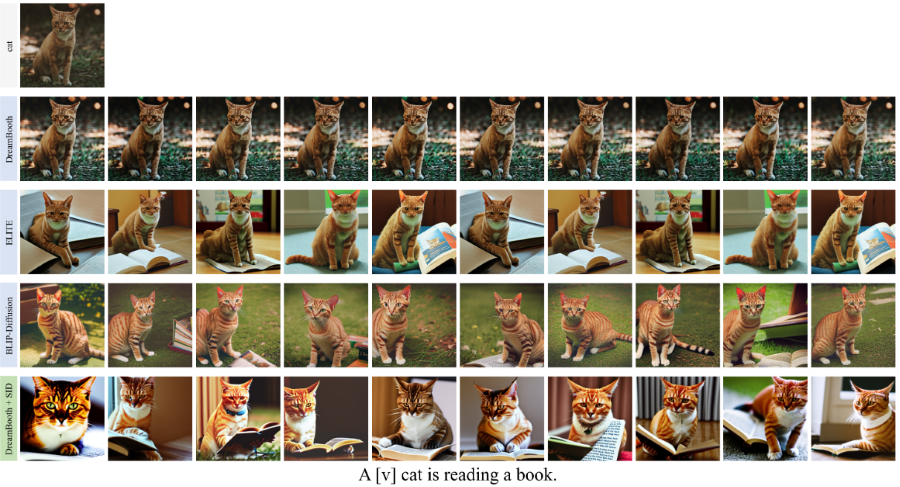}
    \end{subfigure}
    \caption{
    (f) \textbf{Enhancement by SID for a single reference image -- additional example \#6.} 
    When trained with a single reference image, DreamBooth sometimes produces results that are completely overfit to the reference image, particularly when faced with challenging generation prompts.
    % When presented with a challenging generation prompt, DreamBooth sometimes produces results that are completely overfit to the reference image.
    }
\end{figure*}

\begin{figure*}[t] \ContinuedFloat
    \centering
    \begin{subfigure}{\textwidth}
    \includegraphics[width=\textwidth]{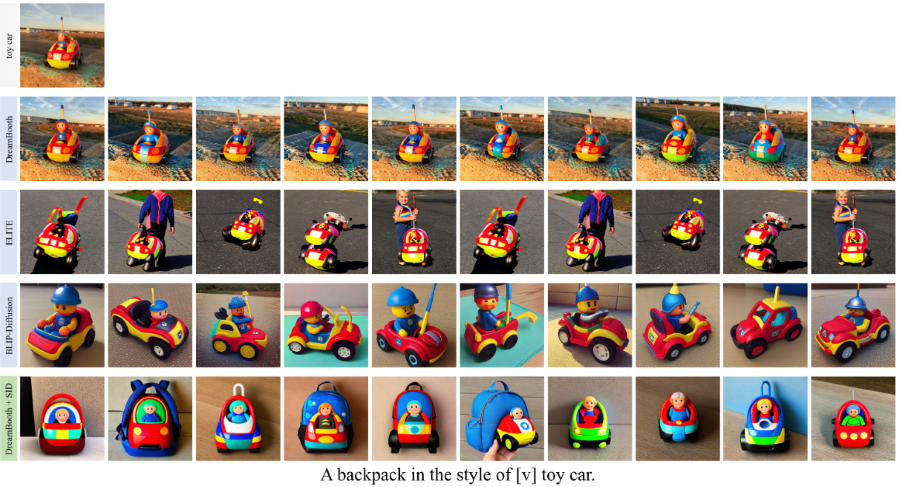}
    \end{subfigure}
    \caption{
    (g) \textbf{Enhancement by SID for a single reference image -- additional example \#7.} 
    While other models failed to generate a backpack, the SID-integrated DreamBooth successfully produced a backpack in the style of the main subject.
    }
\end{figure*}

\begin{figure*}[t] \ContinuedFloat
    \centering
    \begin{subfigure}{\textwidth}
    \includegraphics[width=\textwidth]{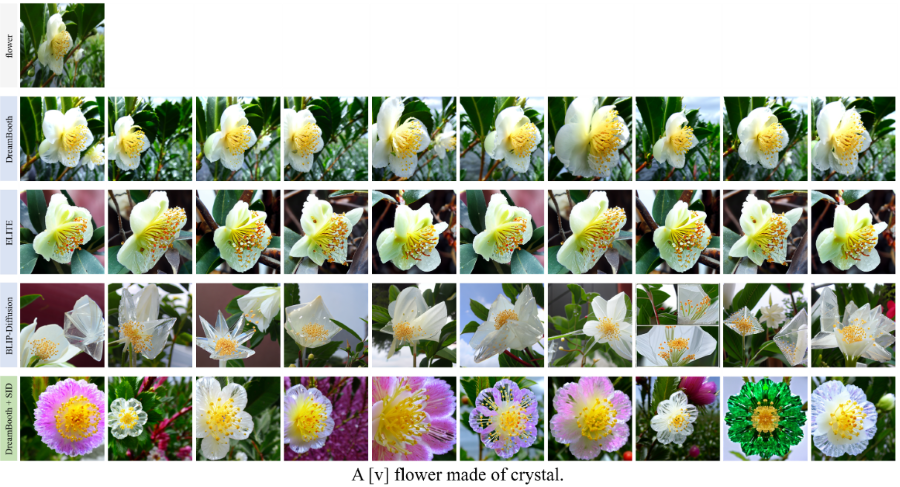}
    \end{subfigure}
    \caption{
    (h) \textbf{Enhancement by SID for a single reference image -- additional example \#8.} 
    SID-integrated DreamBooth successfully altered the material of the main subject while preserving its identity.
    }
\end{figure*}

% \begin{figure*}[t]
%     \centering
%     \begin{subfigure}{\textwidth}
%     \includegraphics[width=\textwidth]{Figures/Appendix/Multi_vs_Neg/Multi_vs_Neg_1.png}
%     \end{subfigure}
%     \caption{(a)}
% \end{figure*}

% \begin{figure*}[t]
%     \centering
%     \begin{subfigure}{\textwidth}
%     \includegraphics[width=\textwidth]{Figures/Appendix/Multi_vs_Neg/Multi_vs_Neg_1.png}
%     \end{subfigure}
%     \caption{(a)}
% \end{figure*}

% \begin{figure*}[t]
%     \centering
%     \begin{subfigure}{\textwidth}
%     \includegraphics[width=\textwidth]{Figures/Appendix/Multi_vs_Neg/Multi_vs_Neg_1.png}
%     \end{subfigure}
%     \caption{(a)}
% \end{figure*}

\begin{figure*}[t]
    \centering
    \begin{subfigure}{\textwidth}
    \centering
    \includegraphics[width=0.9\textwidth]{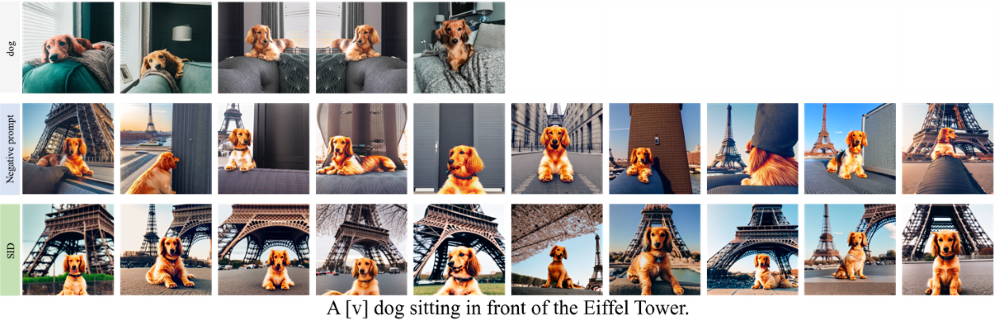}
    \vspace{0.2cm}
    
    \includegraphics[width=0.9\textwidth]{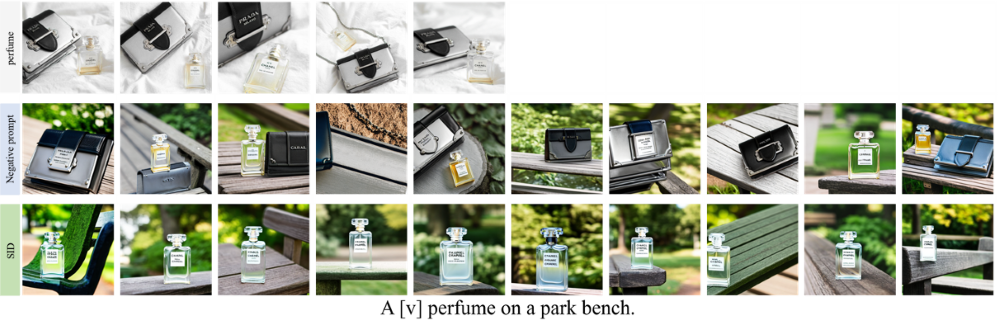}
    \vspace{0.2cm}
    
    \includegraphics[width=0.9\textwidth]{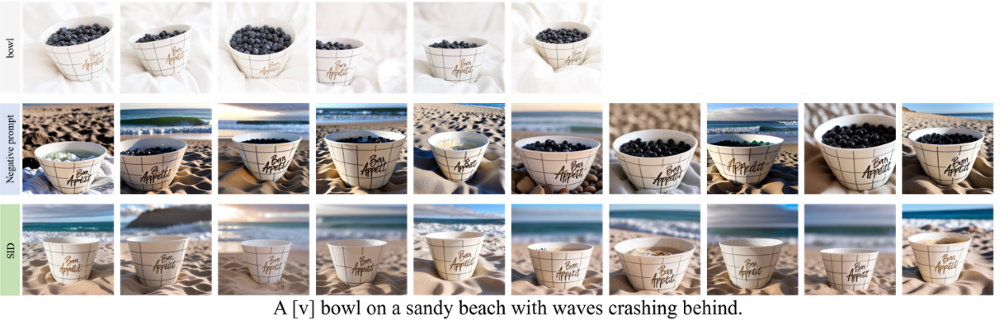}
    \vspace{0.2cm}
    
    \includegraphics[width=0.9\textwidth]{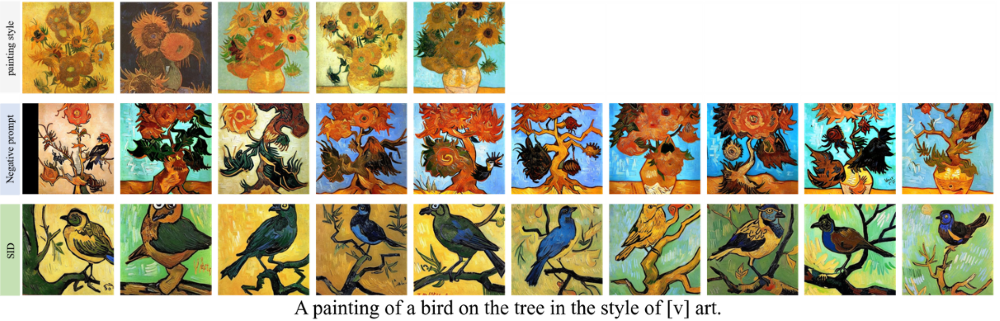}
    \end{subfigure}
    \caption{(a) \textbf{Comparison with negative prompt -- additional examples.} Even when a negative prompt is employed, it appears challenging to counter the effects of undesired embedding entanglements.
    Negative prompts used: ``sitting on the fluffy blanket and green couch" (1st row), ``next to a black Prada purse with silver hardware on a bed of white sheets" (2nd row), ``full of blueberries is on the bed of white sheet" (3rd row), ``a bouquet of sunflowers in the round vase" (last row). DreamBooth is used as the base model. }
    % \label{fig:vs_Neg_Seg}
    \label{fig:C5 (a)}
\end{figure*}

\begin{figure*}[t]\ContinuedFloat
    \centering
    \begin{subfigure}{\textwidth}
    \centering
    \includegraphics[width=0.9\textwidth]{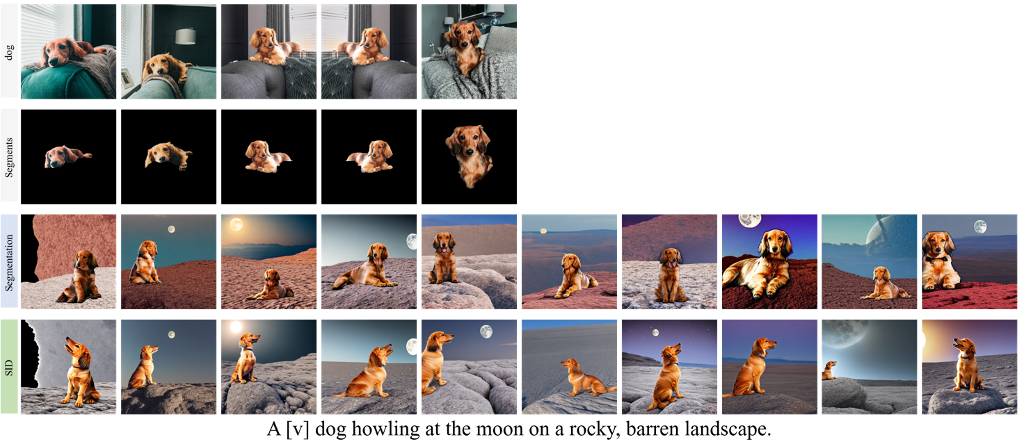}
    \vspace{0.2cm}
    
    \includegraphics[width=0.9\textwidth]{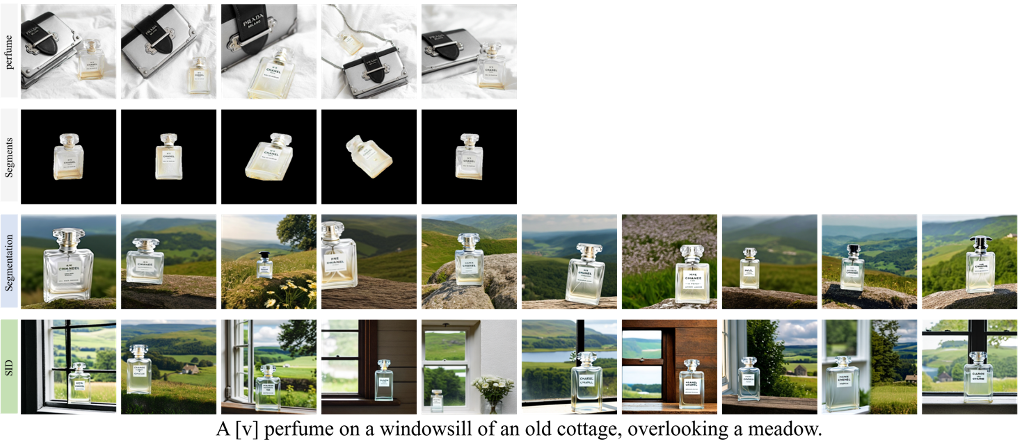}
    \vspace{0.2cm}
    
    \includegraphics[width=0.9\textwidth]{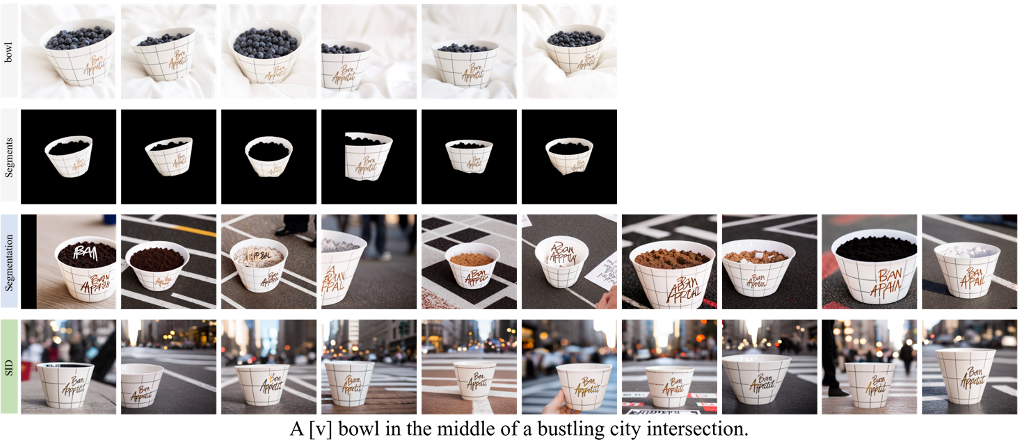}
    \label{fig:C5 (b)}
    \end{subfigure}
    \caption{(b) \textbf{Comparison with segmentation -- additional examples.} 
    Employing segmentation to mitigate undesired embedding entanglement still presents certain limitations.
    The first row highlights the constraint of dynamically changing poses. 
    The second row underscores the incapacity to generate intricate backgrounds, possibly due to the common presence of a black background in segmentation.
    The last row illustrates that removal of tied objects (blueberries in this case) may lead to generated images with peculiar artifacts. DreamBooth is used as the base model.
    }
\end{figure*}
\begin{figure*}[t]
    \begin{subfigure}{\textwidth}
    \centering
    \includegraphics[width=0.9\textwidth]{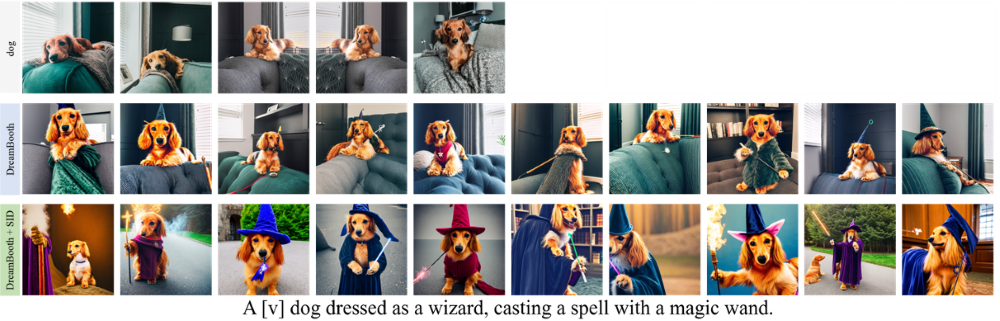}
    \vspace{0.2cm}

    \includegraphics[width=0.9\textwidth]{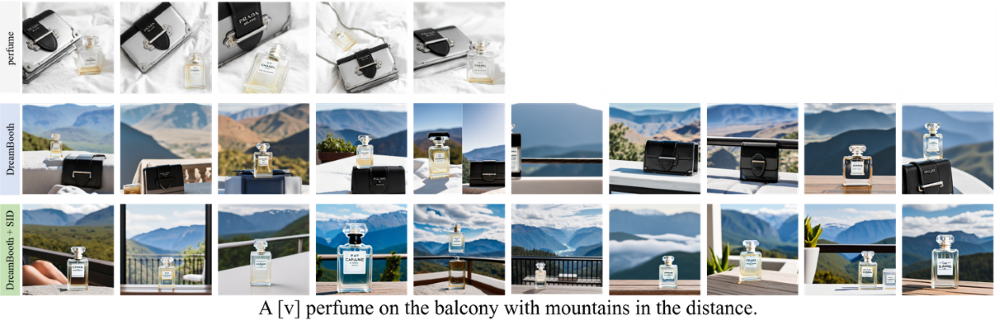}
    \vspace{0.2cm}

    \includegraphics[width=0.9\textwidth]{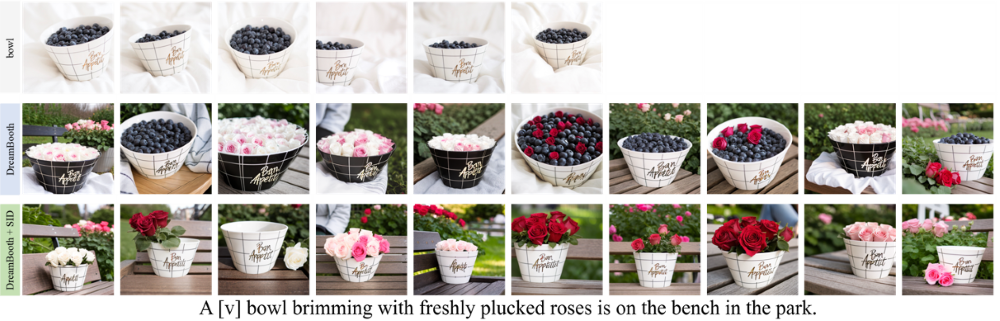}
    \vspace{0.2cm}

    \includegraphics[width=0.9\textwidth]{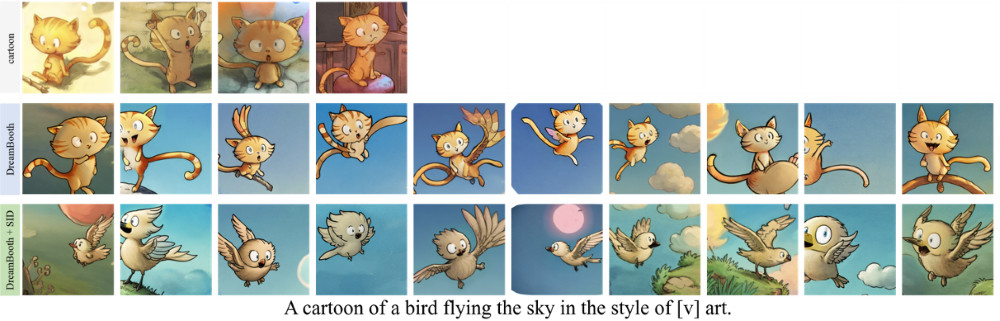}
    \end{subfigure}
    \caption{(a) \textbf{Comparing DreamBooth with its SID-integration in highly-biased scenarios.}
    DreamBooth suffers from undesired embedding entanglement represented by indoor background (1st row), nearby purse (2nd row),
     filled-in blueberries (3rd row), and cat substance (last row).
    % filled-in blueberries (3rd row), 
    % and standing bear substance (last row). 
    SID-integration is definitely required for high-quality text-to-image personalization.
    Image credit: \href{https://www.peppercarrot.com/}{David Revoy} (last row).}
    \label{fig:C6 (a)}
\end{figure*}

\begin{figure*}[t] \ContinuedFloat
    \begin{subfigure}{\textwidth}
    \centering
    \includegraphics[width=0.9\textwidth]{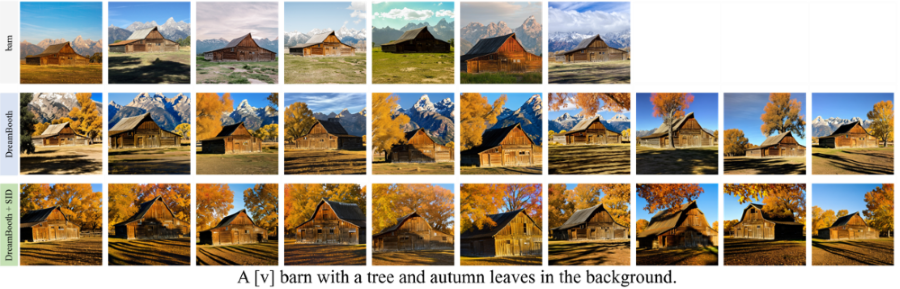}
    \vspace{0.2cm}

    \includegraphics[width=0.9\textwidth]{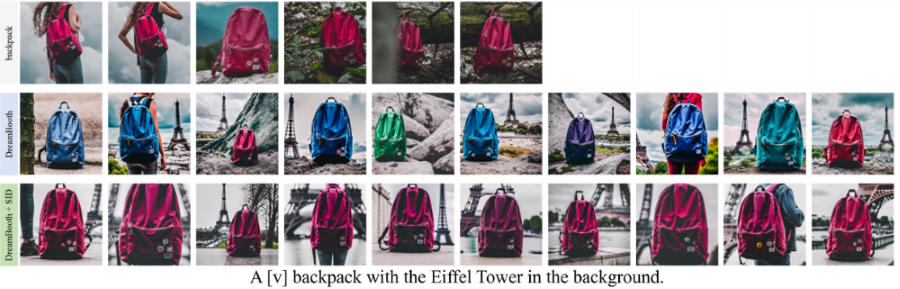}
    \vspace{0.2cm}

    \includegraphics[width=0.9\textwidth]{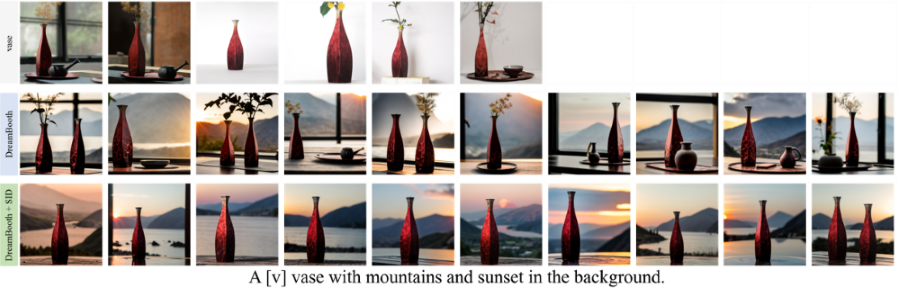}
    \vspace{0.2cm}

    \includegraphics[width=0.9\textwidth]{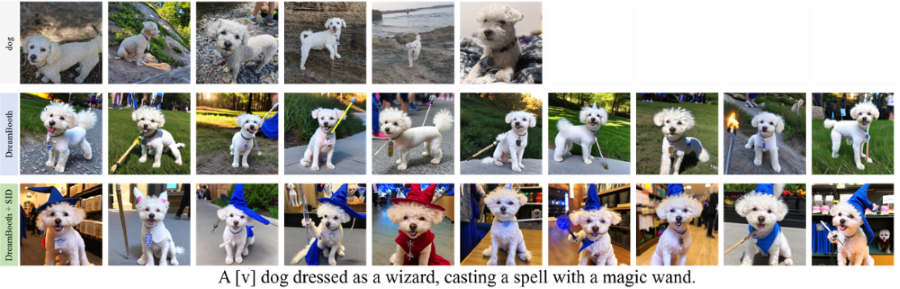}
    \end{subfigure}
    \caption{
    (b) \textbf{Comparing DreamBooth with its SID-integration in moderately-biased scenarios.}
    DreamBooth still suffers from undesired embedding entanglement even in moderately-biased scenarios. 
    This is evident in the background mountain (1st row), rocky surface (2nd row), nearby teapot, saucer, and filled-in plants (3rd row), and grassy field (last row).
    In particular, in the second row, DreamBooth encounters difficulty preserving the identity of the subject, as it tends to change its color. SID-integration is definitely required for high-quality text-to-image personalization.
    }
\end{figure*}

\begin{figure*}[t] \ContinuedFloat
    \begin{subfigure}{\textwidth}
    \centering
    \includegraphics[width=0.9\textwidth]{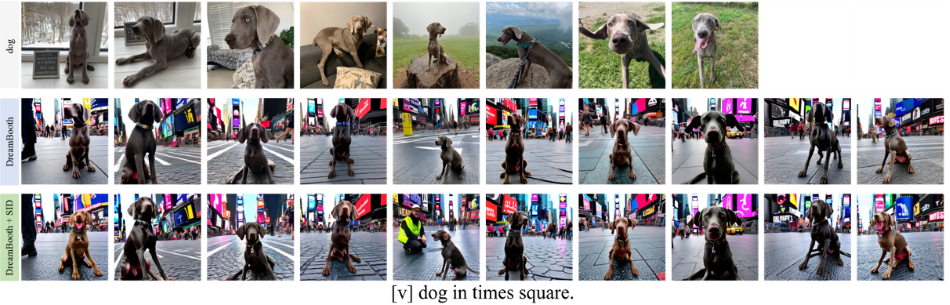}
    \vspace{0.2cm}

    \includegraphics[width=0.9\textwidth]{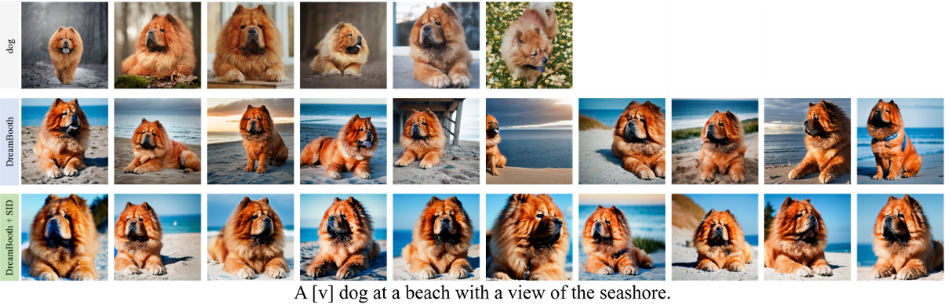}
    \vspace{0.2cm}

    \includegraphics[width=0.9\textwidth]{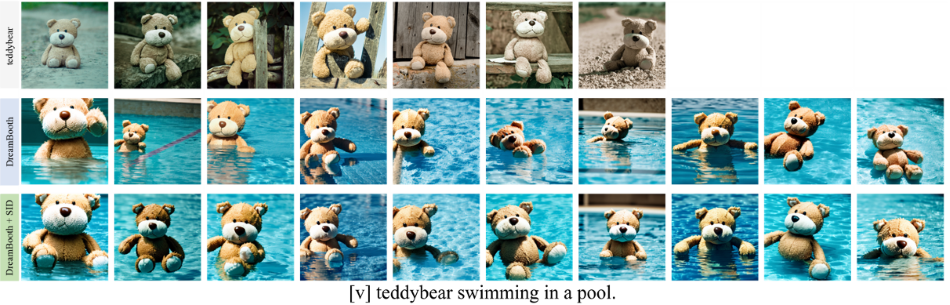}
    \vspace{0.2cm}

    \includegraphics[width=0.9\textwidth]{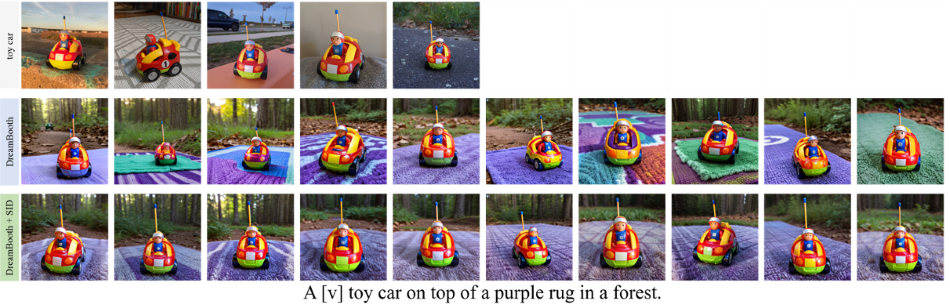}
    \end{subfigure}
    \caption{
    (c) \textbf{Comparing DreamBooth with its SID-integration in low-biased scenarios.} In scenarios with much less or almost no bias in the reference images, both DreamBooth and SID-integrated DreamBooth demonstrate remarkable performance.
    }
\end{figure*}
% \vspace{-0.3cm}
\vspace{-0.1cm}
\section{Societal impact}
Our approach enhances personalized image synthesis, making it easier to create realistic images of personalized subjects in new contexts, even with highly biased reference images, including just a single reference image. While this advancement fosters creativity and contributes to the sharing of personalized content that closely aligns with user guidance, it also raises concerns about potential misuse by malicious users who may exploit generative models for deception or unauthorized copyright infringement. Additionally, these generative models inherit biases from the large-scale dataset used in the pre-training stage, which could inadvertently misinform the public. Future research should prioritize defending against such misuse and reducing biases in generative models to ensure responsible and ethical use, particularly in personalized image synthesis.

% % C
% \input{Suppl_TEX/831.Tab_Case_quantitative}
% \input{Suppl_TEX/931_1.Prompt_cases}
% \input{Suppl_TEX/931_2.Prompt_cases_attentionmap}

% \input{Suppl_TEX/832.Tab_VLM_quantitative}
% \input{Suppl_TEX/932.VLMs}

% \input{Suppl_TEX/933_1.Multi_vs_others}
% \input{Suppl_TEX/934_1.One_vs_encoder}
% \input{Suppl_TEX/935.Multi_vs_Neg_Seg}
% \input{Suppl_TEX/936.Multi_vs_DB}
% % 

% % D
% \input{Suppl_TEX/941.Failure_cases}
% \input{Suppl_TEX/942.Facial_expression}
% % 

% \printbibliography[heading=subbibliography]
% \end{refsection}

% {\small
% \bibliographystylesupple{ieee_fullname}
% \bibliographysupple{egbib}
% }
% \bibliographystylesupple{ieee_fullname}
% \bibliographysupple{egbib_supp}
% {\small
% \bibliographystylesupple{ieee_fullname}
% \bibliographysupple{egbib}
% }

\end{document}